\documentclass{article}

\usepackage{arxiv}

\usepackage[utf8]{inputenc} 
\usepackage[T1]{fontenc}    
\usepackage{hyperref}       
\usepackage{url}            
\usepackage{booktabs}       
\usepackage{amsfonts}       
\usepackage{nicefrac}       
\usepackage{microtype}      
\usepackage{amsmath} 

\usepackage{cleveref}       
\usepackage{lipsum}         
\usepackage{graphicx}
\usepackage{natbib}
\usepackage{doi}

\usepackage{comment}
\usepackage{listings}
\usepackage{booktabs}
\usepackage{multirow}
\usepackage{amsfonts}
\usepackage{pifont} 
\usepackage{colortbl} 
\usepackage{tabularray}
\usepackage{titletoc}
\usepackage{amssymb}

\usepackage{tcolorbox}
\tcbuselibrary{breakable}
\usepackage{caption}

\usepackage{titletoc} 

\usepackage{url} 

\newenvironment{links}{%
  \begin{center}
  \begin{minipage}{0.85\linewidth} 
  \begin{list}{}{\setlength\itemsep{0pt}\setlength\leftmargin{0pt}}
}{%
  \end{list}
  \end{minipage}
  \end{center}
}
\newcommand{\link}[2]{\item \textbf{#1:} \url{#2}}

\title{MiDAS: A Multimodal Data Acquisition System\\ and Dataset for Robot-Assisted\\ Minimally Invasive Surgery}

\usepackage{authblk}

\newbox{\orcid}\sbox{\orcid}{\includegraphics[scale=0.06]{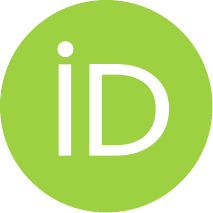}} 

\author[1]{Keshara Weerasinghe}
\author[1]{Seyed Hamid Reza Roodabeh}
\author[2]{Andrew Hawkins (MD)}
\author[1]{\\Zhaomeng Zhang}
\author[1]{Zachary Schrader}
\author[1]{Homa Alemzadeh}

\affil[1]{School of Engineering and Applied Science, University of Virginia}
\affil[2]{Thoracic and Cardiovascular Surgery, Department of Surgery, University of Virginia Health System}
\affil[1]{\texttt{\{cjh9fw, ydq9ag,  sus7sv, zgr3jk, ha4d\}@virginia.edu}}
\affil[2]{\texttt{adh2tp@uvahealth.org}}

\renewcommand{\shorttitle}{}

\hypersetup{
    pdftitle={MiDAS: A Multimodal Data Acquisition System and Dataset for Robot Assisted Minimally Invasive Surgery},
    pdfsubject={cs.CV, cs.HC, cs.AI},
    pdfauthor={Keshara Weerasinghe, Seyed Hamid Reza Roodabeh, Andrew Hawkins (MD), Zhaomeng Zhang, Zachary Schrader, Homa Alemzadeh},
    pdfkeywords={Robot assisted surgery, RMIS, Surgical Activity Recognition, Multimodal dataset, Open source data collection system, Action recognition, Suturing Dataset, Peg Transfer dataset}
}

\usepackage{chngcntr}

\newcommand{\appendixformat}{
  \appendix
  \renewcommand{\thesection}{\Alph{section}}
  \renewcommand{\thesubsection}{\thesection.\arabic{subsection}}
  \renewcommand{\thesubsubsection}{\thesubsection.\arabic{subsubsection}}
}

\begin{document}
\maketitle

\begin{abstract}

\textbf{Background:} Robot-Assisted Minimally Invasive Surgery (RAMIS) research increasingly relies on multimodal data, yet access to proprietary robot telemetry remains a barrier. We introduce the Multimodal Data Acquisition System (MiDAS), an open-source, platform-agnostic system enabling time-synchronized, non-invasive multimodal data acquisition across surgical robotic platforms. 

\textbf{Methods:} MiDAS integrates electromagnetic and depth camera-based hand tracking, foot pedal sensing, and surgical video capturing without requiring proprietary robot interfaces. We validated MiDAS on the open-source Raven-II and the clinical da Vinci Xi by collecting multimodal datasets of Peg Transfer and Hernia Repair Suturing tasks performed by surgical residents. Correlation analysis and gesture recognition experiments were conducted. 

\textbf{Results:} External hand and foot sensing closely approximated internal robot kinematics and non-invasive motion signals achieved gesture recognition performance comparable to proprietary telemetry. 

\textbf{Conclusions:} MiDAS enables reproducible multimodal RAMIS data collection and is released with annotated datasets, including the first multimodal dataset of hernia repair suturing on simulation models.

\end{abstract}

\begin{links}
    \link{Code \& Dataset}{https://uva-dsa.github.io/MiDAS/}
\end{links}




\section{Introduction}\label{intro}

Robot-Assisted Minimally Invasive Surgery (RAMIS) has become central to modern surgical procedures, particularly in minimally invasive settings~\cite{nadeem2025robot},
yet surgical outcomes remain heavily dependent on surgeon skill~\cite{Wang2018}. The multimodal data generated during RAMIS, including stereo video, robot kinematics, and interaction events offer unprecedented opportunities for training assessment, error detection and skill transfer research~\cite{lukacs2026scars, yasar2020real,hung2018automated}.

At the system level, the surgeon operates from a master console, viewing the surgical field through an Endoscope Camera Module (ECM) that provides stereoscopic vision, and controls the surgical instruments via Master Tool Manipulators (MTMs). The MTMs translate the surgeon's hand motion to low-level commands that drive the Patient Side Manipulators (PSMs), where forward and inverse kinematics close the control loop to actuate the instruments. This process generates rich low-level telemetry, such as joint states and end-effector poses, that complement the surgical video feed. These multimodal signals are critical for activity recognition~\cite{huaulme2023peg} and safety monitoring and error mitigation~\cite{yasar2020real, actvitiy_aware_recovery}, particularly when visual data are compromised by occlusion or lens contamination~\cite{allers2016evaluation}.

Although Intuitive Surgical’s da Vinci systems remain the most widely used robotic surgical platform in clinical practice, the ecosystem is expanding. Medtronic’s Hugo platform received European approval for urology and gynecology procedures in 2021~\cite{cepolina2022introductory}, and Medicaroid Corporation's Hinotori surgical robot has been employed for actual surgeries~\cite{MIYAMOTO2025162}. Research groups routinely employ open-source and industrial robots adapted for surgical tasks, including Raven-II~\cite{hannaford2012raven}, Taurus II, and ABB’s YuMi~\cite{madapana2019desk}. Simulation environments are also prevalent, such as SimNow (Intuitive) and virtual models of Raven-II~\cite{li2016hardware} and Taurus II~\cite{madapana2019desk}.

However, access to time-synchronized, multimodal RAMIS data remains limited by proprietary systems, institutional privacy constraints, and restricted access to clinical platforms. Open research systems with built-in data collection capabilities, such as Raven-II~\cite{hannaford2012raven} and dVRK (based on decommissioned second/third generation da Vinci S/Si robots)~\cite{kazanzides2014open,xu2025dvrk}, mitigate these barriers to some extent, but still require specialized infrastructure and do not represent the current generation of commercial surgical robots. Furthermore, most large-scale datasets from RAMIS are vision-only~\cite{psychogyios2023sar,huaulme2021micro} and the existing multimodal datasets combining robot kinematics and video have primarily focused on simple dry-lab training tasks~\cite{madapana2019desk,gao2014jhu}.

\begin{figure*}[htb]
\centerline{\includegraphics[width=\linewidth]{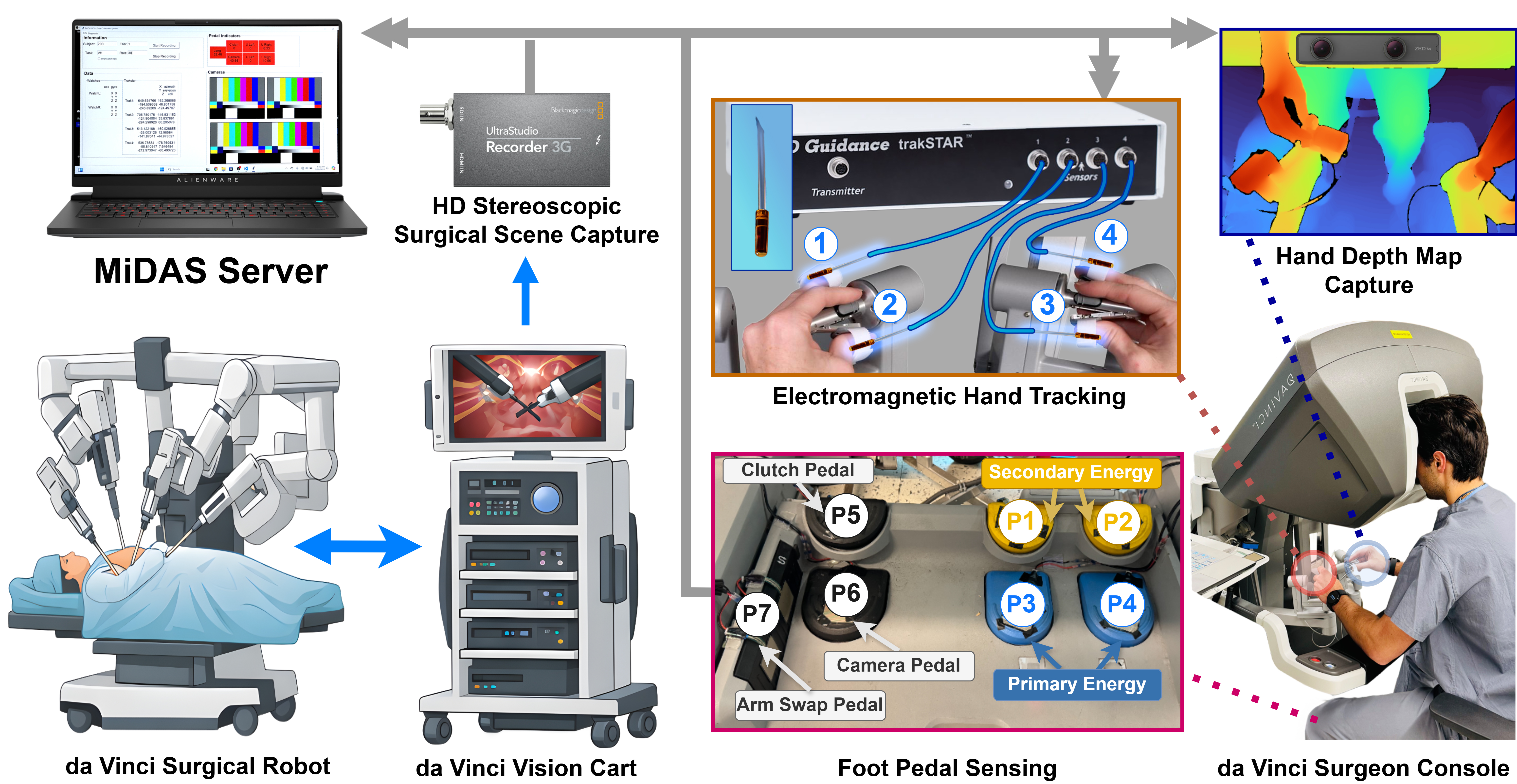}}
\caption{The Data Collection System deployed on a da Vinci Surgical Xi system~\cite{daVinciimage}.\label{main}}
\end{figure*}

To address the scarcity of kinematics telemetry in RAMIS, different alternative modalities have been explored.~\cite{hashemi2023acquisition} employed a depth-camera system to track surgeon hand motion as MTM proxies while extracting pedal states from graphical overlays. Although eliminating some internal dependencies, purely visual tracking remains sensitive to occlusion and line-of-sight constraints. 
In parallel, prior work has evaluated commercial hand-tracking systems in surgical contexts. Optical sensors such as Leap Motion and Kinect demonstrated insufficient precision and dynamic robustness for surgical interfaces~\cite{guna2014analysis,kim2014experimental}. More recent studies using electromagnetic tracking~\cite{goldbraikh2022using,goldbraikh2023kinematic} achieved strong gesture recognition performance in dry-lab suturing settings. However, systematic validation of externally sensed trajectories against ground-truth robot kinematics across platforms remains limited. Building on findings regarding electromagnetic tracking for surgical robot kinematics~\cite{KayDissertation}, alternative external sensing modalities offer a promising direction for platform-agnostic acquisition without requiring direct telemetry access.

Several external data acquisition systems have been also developed to integrate with the existing surgical robotic platforms.~\cite{10585836} introduced a multimodal collection framework for the dVRK that captures MTM, PSM, and ECM data through joint states, forward kinematics, fiducial-based optical pose estimation, and hand–eye calibration, along with pedal signals obtained from internal message streams and electrosurgical unit outputs. The collected data, derived from practice cholecystectomy procedures, were evaluated for tasks such as pedal press prediction and object segmentation. While the dataset contains multimodal data, it does not include gesture- or activity-level annotations to support surgical workflow modeling or activity recognition tasks.
A more recent work introduced a multimodal data collection framework and wet-lab dataset for imitation learning in robotic cholecystectomy, comprising of endoscopic video and internally recorded PSM and MTM kinematics data from a dVRK platform~\cite{hansen2026imitatecholec}, complemented by instrument-mounted wrist cameras for improved depth perception. Both frameworks rely on platform-specific internal data streams, limiting their generalizability to other systems and clinical settings.

Beyond data acquisition systems, several public surgical activity datasets have shaped benchmarking in robotic surgery. DESK~\cite{madapana2019desk} provides dry-lab Peg Transfer data with video and PSM kinematics, while JIGSAWS~\cite{gao2014jhu} remains a foundational dry-lab benchmark for suturing, knot-tying, and needle-passing with synchronized video and robot kinematics from dVRK. More clinically realistic datasets have also emerged. SAR-RARP50~\cite{psychogyios2023sar} provides in-vivo suturing annotations from robotic prostatectomy, although it is limited to vision data. ImitateCholec~\cite{hansen2026imitatecholec} introduces multimodal ex vivo cholecystectomy data with internally recorded dVRK kinematics and video. While a positive step, the resulting dataset, primarily aimed at imitation learning research, was collected using only a single unit of dVRK from short trials of clipping and cutting actions performed by two non-clinical research assistants. Table~\ref{tab:dataset_stats_combined} shows a comparative presentation of these datasets in terms of task setting, sensing modalities, annotation granularity, sample counts, and duration.

Collectively, these prior datasets and data collection systems highlight important but complementary limitations: dry-lab benchmarks provide structured gesture labels but limited procedural realism, in-vivo video datasets capture realistic surgical workflows but often lack synchronized kinematic and interaction signals, and recent multimodal wet-lab datasets remain tied to specific robotic platforms and internal telemetry streams. Moreover, systematic validation of external sensing modalities against ground-truth robot kinematics remains limited. These gaps motivate the development of a non-invasive, platform-agnostic solution for multimodal data acquisition.

To address these challenges, we present an open-source, platform-agnostic Multimodal Data Acquisition System (MiDAS) that enables time-synchronized, non-invasive data collection across robotic platforms without modifying or interfacing directly with proprietary control systems (Figure \ref{main}). MiDAS leverages Electromagnetic Hand Tracking (EmHT) and complementary sensing modalities to approximate end-effector kinematics when direct robot telemetry is inaccessible. We quantify approximation fidelity by benchmarking EmHT-derived trajectories against ground-truth robot kinematics on the Raven-II and demonstrate deployment on a clinical da Vinci Xi system during a robotic surgery training bootcamp held at the University of Virginia Hospital in 2024. 

In summary, our contributions are as follows:
\begin{enumerate}
    \item We present an open-source Multimodal Data Acquisition System (MiDAS) that can be non-invasively integrated with standard teleoperated robotic surgery systems to collect fully synchronized data in real-time.
    \item We propose new alternative modalities to proprietary robot kinematics, including surgeon hand and foot motion kinematics (captured using electromagnetic hand tracking sensors, depth cameras, and force sensors) and demonstrate that they closely approximate surgeon and robot kinematics and can be effectively used for key tasks such as activity recognition. 

    \item We evaluate MiDAS by collecting data from two representative tasks: standard Peg Transfer on Raven-II and a novel multimodal dataset of hernia repair suturing on realistic KindHeart tissue models using the da Vinci Xi, performed during a training bootcamp at the University of Virginia Hospital. We identify effective modality combinations for multimodal gesture recognition by benchmarking state-of-the-art (SOTA) transformer-based~\cite{weerasinghe2024multimodal} and convolutional neural networks (CNN)-based models~\cite{farha2019ms}.
    
    \item We publicly release both the data collection system and the resulting multimodal, multi-platform datasets with gesture annotations, providing the community with a unique testbed for exploring multimodal learning, alternative sensing, and downstream applications such as activity recognition, skill assessment~\cite{lukacs2026scars}, and error detection~\cite{hutchinson2022analysis} in RAMIS.

\end{enumerate}

\begin{table*}[!t]
\centering
\resizebox{\textwidth}{!}{%
\begin{tabular}{@{}llcccccccc@{}}
\toprule
\textbf{Dataset} & \textbf{Tasks} & \multicolumn{4}{c}{\textbf{Data Modalities}} & \multicolumn{4}{c}{\textbf{Annotations}}\\
\cmidrule(lr){3-6} \cmidrule(l){7-10}
 & & \textbf{PSM} & \textbf{MTM} & \textbf{Video} & \textbf{Pedals} & \textbf{ID} & \textbf{Gesture Classes} & \textbf{Samples} & \textbf{Avg. Dur. (s)}    \\
\midrule
\multirow{7}{*}{\shortstack[l]{\textbf{Raven-II}\\\textit{Dry-lab}\\(15 trials)}}
 & \multirow{7}{*}{Peg Transfer} 
 & \multirow{7}{*}{$\checkmark$} 
 & \multirow{7}{*}{$\checkmark$} 
 & \multirow{7}{*}{$\checkmark$} 
 & \multirow{7}{*}{$\checkmark$} 
 & S1 & Approach peg                 & 49  & $5.84 \pm 3.66$ \\
 & &  &  &  &  & S2 & Align \& grasp               & 50  & $6.05 \pm 7.22$ \\
 & &  &  &  &  & S3 & Lift peg                     & 51  & $2.83 \pm 1.28$ \\
 & &  &  &  &  & S4 & Transfer peg -- Get together & 52  & $8.08 \pm 4.04$ \\
 & &  &  &  &  & S5 & Transfer peg -- Exchange     & 50  & $9.65 \pm 6.85$ \\
 & &  &  &  &  & S6 & Approach pole                & 47  & $5.87 \pm 2.52$ \\
 & &  &  &  &  & S7 & Align \& place               & 46  & $5.37 \pm 2.74$ \\
\cmidrule(lr){7-10}
 & &  &  &  &  &  & \textbf{Total: 7} & \textbf{345} & \textbf{36 (min)} \\
 & &  &  &  &  &  &  &  & $\sim$2.4 mins/trial \\

\midrule

\multirow{8}{*}{\shortstack[l]{\textbf{da Vinci Xi}\vspace{0.15em}\\\textit{High-fidelity }\\\textit{Hybrid Porcine}\\\textit{Tissue, ex vivo}\\(17 trials)}}
 & \multirow{8}{*}{\shortstack[l]{Suturing\\(Inguinal and\\Ventral Hernia\\Repair)}}
 & \multirow{8}{*}{$\dagger$}      
 & \multirow{8}{*}{$\dagger$}      
 & \multirow{8}{*}{$\checkmark$}  
 & \multirow{8}{*}{$\checkmark$}  
 & G1 & Orient Needle                    & 246 & $9.30 \pm 9.48$ \\
 & &  &  &  &  & G2 & Target Needle                    & 321 & $3.61 \pm 7.63$ \\
 & &  &  &  &  & G3 & Push Needle Through Tissue       & 328 & $7.82 \pm 10.88$ \\
 & &  &  &  &  & G4 & Pull Needle out of Tissue        & 273 & $5.72 \pm 6.67$ \\
 & &  &  &  &  & G5 & Reach Suture                     & 39  & $6.10 \pm 6.56$ \\
 & &  &  &  &  & G6 & Pull Suture                      & 233 & $13.12 \pm 21.98$  \\
 & &  &  &  &  & G7 & Make C Loop                      & 143 & $5.07 \pm 3.74$ \\
 & &  &  &  &  & G8 & Square Knot and Cinch            & 141 & $8.47 \pm 6.88$ \\
\cmidrule(lr){7-10}
 & &  &  &  &  &  & \textbf{Total: 8} & \textbf{1,724} &  \textbf{212 (min)}\\
 & &  &  &  &  &  &  &  & $\sim$12.5 min/trial \\
 
\midrule

\multirow{3}{*}{\shortstack[l]{\textbf{DESK}~\cite{madapana2019desk}\\\textit{Dry-lab}\\(76 trials, 228 transfers)}}
 & \multirow{3}{*}{Peg Transfer}
 & \multirow{3}{*}{$\checkmark$} & \multirow{3}{*}{$\times$} & \multirow{3}{*}{$\checkmark$} & \multirow{3}{*}{$\times$}
 &  &  &  &  \\
 &  &  &  &  &  &  & \textbf{Total: 7} & 1,611  & {\textbf{$\sim$65 (min)}}   \\  
 &  &  &  &  &  &  &  &  &  $\sim$0.8 min/trial \\  

\midrule

\multirow{3}{*}{\shortstack[l]{\textbf{JIGSAWS}~\cite{gao2014jhu}\\\textit{Dry-lab}\\(103$^*$ trials)}}
 & \multirow{3}{*}{\shortstack[l]{Suturing (39)\\Knot-Tying (28)\\Needle-Passing (36)}}
 & \multirow{3}{*}{$\checkmark$} 
 & \multirow{3}{*}{$\checkmark$} 
 & \multirow{3}{*}{$\checkmark$} 
 & \multirow{3}{*}{$\times$}
 &  &  & &\\
 &  &  &  &  &  &  & \textbf{Total: 11 } &  1,703 &  {\textbf{$\sim$206 (min)}}  \\   
 &  &  &  &  &  &  &  &  &  $\sim$2 min/trial \\  

\midrule

\multirow{3}{*}{\shortstack[l]{\textbf{SAR-RARP50}~\cite{psychogyios2023sar}\\\textit{Human,  in vivo}\\(50 trials)}}
 & \multirow{3}{*}{\shortstack[l]{Suturing\\(Radical\\Prostatectomy)}}
 & \multirow{3}{*}{$\times$} 
 & \multirow{3}{*}{$\times$} 
 & \multirow{3}{*}{$\checkmark$} 
 & \multirow{3}{*}{$\times$}
 &  &  & &  \\
 &  &  &  &  &  &  & \textbf{Total: 8} &  2,219  & {\textbf{$\sim$267 (min)}} \\   
 &  &  &  &  &  &  &  &  &  $\sim$5.34 min/trial \\  
\midrule

\multirow{3}{*}{\shortstack[l]{\textbf{ImitateCholec}~\cite{hansen2026imitatecholec}\\\textit{Porcine, ex vivo}\\(34$^*$$^*$  trials)}}
 & \multirow{3}{*}{\shortstack[l]{Clipping \& Cutting\\(Cholecystectomy)}}
 & \multirow{3}{*}{$\checkmark$} 
 & \multirow{3}{*}{$\checkmark$} 
 & \multirow{3}{*}{$\checkmark$} 
 & \multirow{3}{*}{$\times$}
 &  &  & &  \\
 &  &  &  &  &  &  & \textbf{Total: 17 ***} &  18,343  & {\textbf{$\sim$1200 (min)}} \\   
 &  &  &  &  &  &  &  &  & $\sim$1.1 min/trial \\   

\bottomrule
\end{tabular}%
}
\caption{MiDAS datasets across two platforms versus datasets from related work. *One JIGSAWS trial is a single instance of three-throw suturing, tying three knots, or passing a needle through a ring three times. **Assuming a single trial corresponds to one ex vivo tissue. ***Four distinct actions (Grasp, Apply Clip, Go Back, Cut) are repeated to clip and cut left and right tubes. $\dagger$-Available from MiDAS electromagnetic and RGB-D hand tracking.} 
\vspace{-1em}
\label{tab:dataset_stats_combined}
\end{table*}

The datasets captured across two distinct platforms (e.g., 15 Peg Transfer trials on Raven-II and 17 Suturing trials on the da Vinci Xi) provide high-fidelity, time-synchronized multimodal signals in realistic surgical training environments. While modest in scale, they serve as a strong initial benchmark for validating the system's versatility across platforms. The primary contribution of this work is the development and validation of the data acquisition system itself, with these datasets laying the groundwork for larger-scale multimodal research.


\section{Materials and Methods}\label{matmethod}

\noindent\textbf{Ethics Statement:} The surgical bootcamp and associated data collection were approved by the University of Virginia Institutional Review Board under protocol IRB-SBS 6724. All participants provided informed consent prior to participation. The study involved surgical residents performing training tasks on simulation models and did not involve patient data.

This section first presents the design, architecture, and implementation of the MiDAS multimodal data acquisition system. We then describe the resulting multimodal datasets collected across two robotic platforms, which serve both to validate the system and to establish reproducible benchmarks for multimodal analysis in robot-assisted surgery.

\subsection{MiDAS: A Multimodal Data Acquisition System}\label{dcs-method}

We present MiDAS, a platform-agnostic, open-source, non-invasive system that records synchronized multimodal data during teleoperated RAMIS to approximate robot kinematic trajectories. Unlike existing setups constrained to proprietary robot telemetry, MiDAS integrates external sensing modules, including EmHT, a depth hand capture camera, and foot pedal force sensors, through a unified client--server software framework. Figure~\ref{main} presents the overall architecture of MiDAS, in which a central server coordinates real-time acquisition from heterogeneous sensing modules, provides a graphical user interface for metadata configuration and system monitoring, and synchronizes the resulting data streams for downstream analysis. This design enables reproducible data collection on both open research platforms (Raven-II) and commercial systems (da Vinci Xi) maintaining platform-agnostic operation. Below, we describe each sensing module and its role in the MiDAS pipeline. Additional architectural and synchronization details are provided in \textit{Appendix A.1}.

\subsubsection{Electromagnetic Hand Tracking}
\label{sec:method:EmHT}

To capture surgeon hand motion non-invasively, we employ the NDI \emph{trakSTAR} electromagnetic tracking device \cite{NDI3DGuidance} with four miniature 6-DoF sensors (3D position and orientation). Two sensors are mounted on each MTM control at the thumb and middle-finger contact pads (Figure \ref{main}), preserving the console hardware while measuring fine manipulator motions. This placement yields continuous finger-level pose trajectories during operation.

MiDAS logs 3D position and orientation for each sensor at 270~Hz. To use these measurements as a proxy for robot kinematics (MTM/PSM \textit{position, orientation, velocity, grasper angle}), raw sensor data are mapped from the electromagnetic tracker frame to the robot reference frames via a calibrated rigid-body transformation (rotation and translation). To account for residual modeling errors, sensor noise, and controller inaccuracies, we augment this transformation with a learned residual modeled using a Multi-Layer Perceptron (MLP). 
We evaluate the accuracy and correlation of the resulting trajectories against ground-truth MTM/PSM poses on the Raven-II platform in Sec.~\ref{emht_handkp_correlation_analysis}, with additional details in \textit{Appendix~A.2}.

\subsubsection{Depth Camera-Based Hand Tracking}

To capture fine-grained hand movements, including manipulating finger positions within the MTM workspace, we mount a Red-Green-Blue-Depth (RGB-D) stereo camera above the surgeon console. Specifically, we use a ZED Mini camera~\cite{StereolabsZEDMini}, which provides synchronized 720p color video and dense depth maps at 30~Hz, enabling markerless 3D reconstruction of hand poses and manipulation trajectories without instrumenting the surgeon. 

To approximate robot end-effector motions, we extract thumb and middle-finger 3D positions for each hand and transform them into the MTM and PSM frames using a lightweight MLP, yielding Hand Key-Point features denoted as \textit{HandKP}. 
Limited camera field-of-view (FoV) and occlusions cause intermittent keypoint detection gaps. We fill gaps of at most one second using cubic spline interpolation and longer gaps using forward-filling. Details on keypoint detection, coordinate frame transformations and the training of the MLP model, and camera calibration are provided in \textit{Appendix A.3}.

\subsubsection{High-Definition Stereoscopic Video Capture}
We use Open Broadcaster Software (OBS)~\cite{OBS} as a unified capture layer to support stereoscopic video recording across platforms. On da~Vinci~Xi, stereoscopic Serial Digital Interface (SDI) output is captured using Blackmagic SDI recorders, while on Raven-II, stereo video is recorded using a ZED camera. OBS records synchronized, high-definition stereo streams at 30~Hz. Recording is controlled programmatically via the OBS WebSocket API, enabling deterministic session control and integration with server.

\subsubsection{Foot Pedal Sensing}

The Foot Pedal Sensing (FPS) captures surgeon foot pedal interactions, which encode essential behavioral cues during RAMIS including instrument activation, clutching, and potential markers of expertise. To detect pedal presses non-invasively, we developed a \textit{compact, low-cost, open-source module} built around an Arduino-class microcontroller and thin-film force-sensitive resistors (FSRs)~\cite{adafruit_fsr_2012}. Sensors are affixed directly to the console pedal surfaces, enabling continuous monitoring without modifying hardware or altering the surgeon’s tactile experience.

FPS supports up to nine pedals and records both binary pedal states and timestamped analog signals at 30~Hz, providing detailed activation traces for downstream analysis. During initial setup, each sensor is mapped to the corresponding robot pedal, followed by a brief calibration to determine pedal-specific thresholds for reliable actuation detection. These thresholds are further refined using a data-driven procedure based on ground-truth pedal labels. The complete hardware design, calibration steps, signal-processing details, and threshold-optimization methodology are provided in \textit{Appendix A.4}.

\begin{figure*}[!t]
    \centering
    
    \begin{minipage}[t]{0.48\textwidth}
        \centering
        \includegraphics[width=\linewidth]{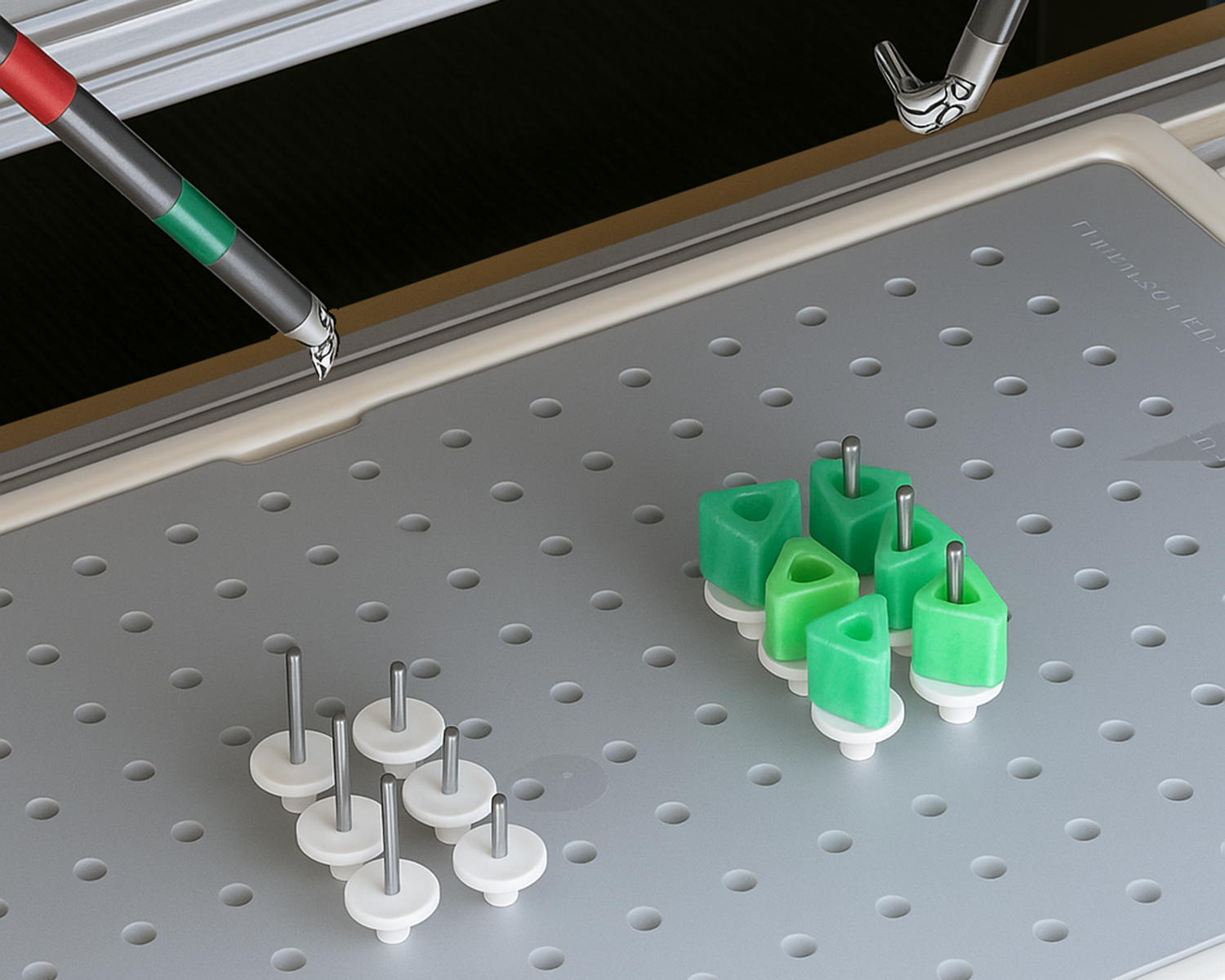}\\[3pt]
        \textbf{(a)} Raven-II Peg Transfer Task
    \end{minipage}
    \hfill
    \begin{minipage}[t]{0.48\textwidth}
        \centering
        \includegraphics[width=\linewidth]{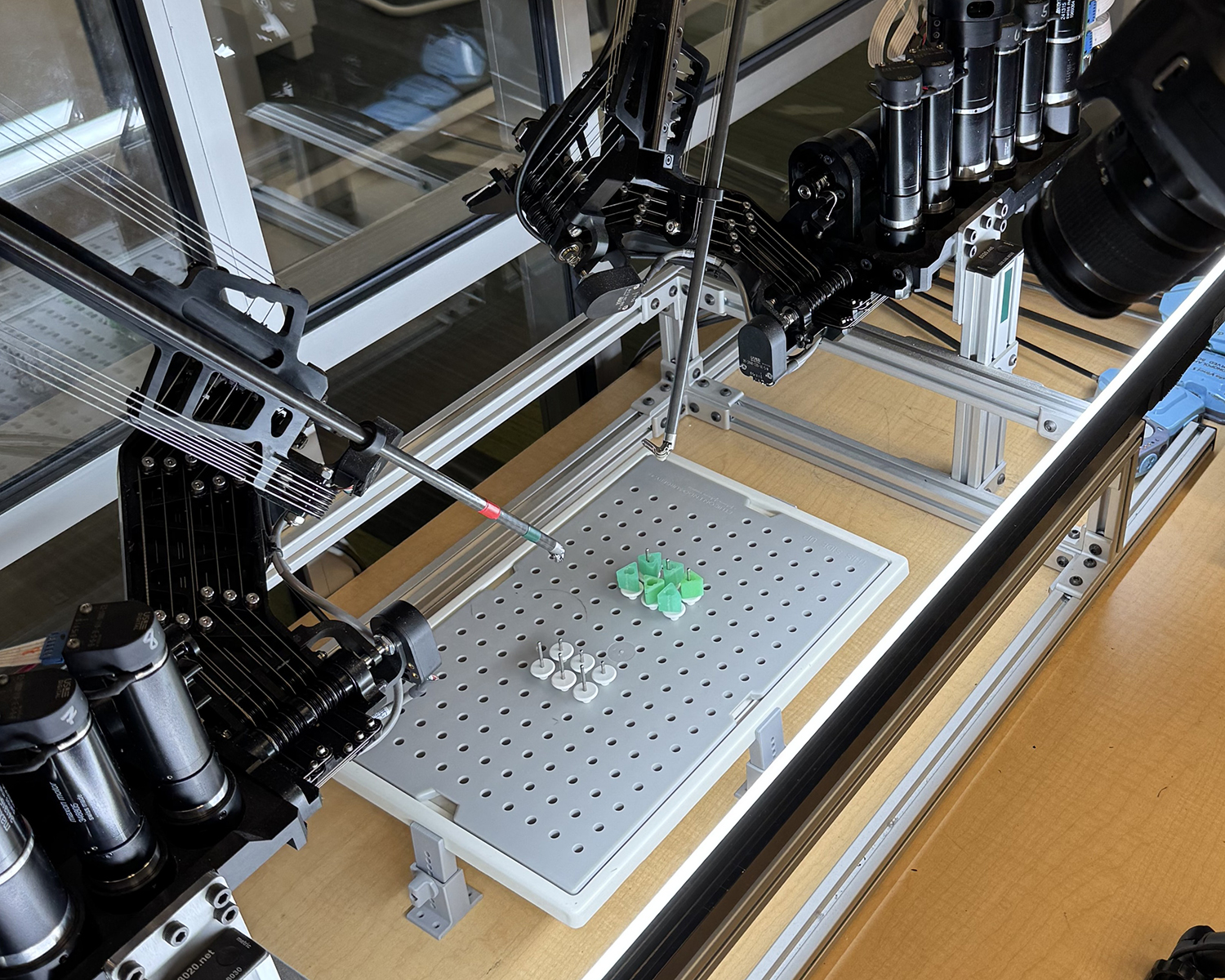}\\[3pt]
        \textbf{(b)} Raven-II Dry-Lab Setup
    \end{minipage}

    \vspace{5pt}

    \begin{minipage}[t]{0.48\textwidth}
        \centering
        \includegraphics[width=\linewidth]{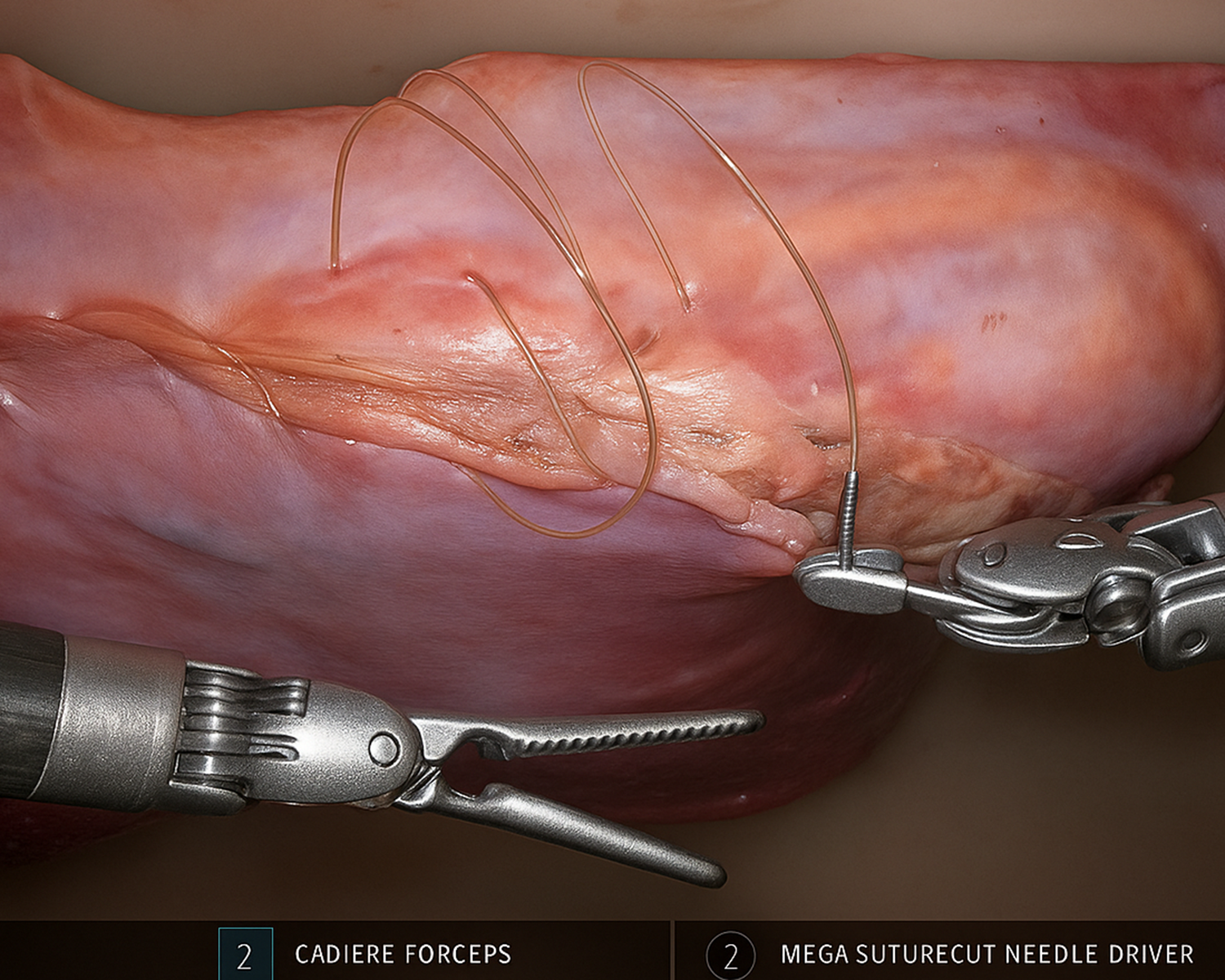}\\[3pt]
        \textbf{(c)} da Vinci Xi Suturing Task
    \end{minipage}
    \hfill
    \begin{minipage}[t]{0.48\textwidth}
        \centering
        \includegraphics[width=\linewidth]{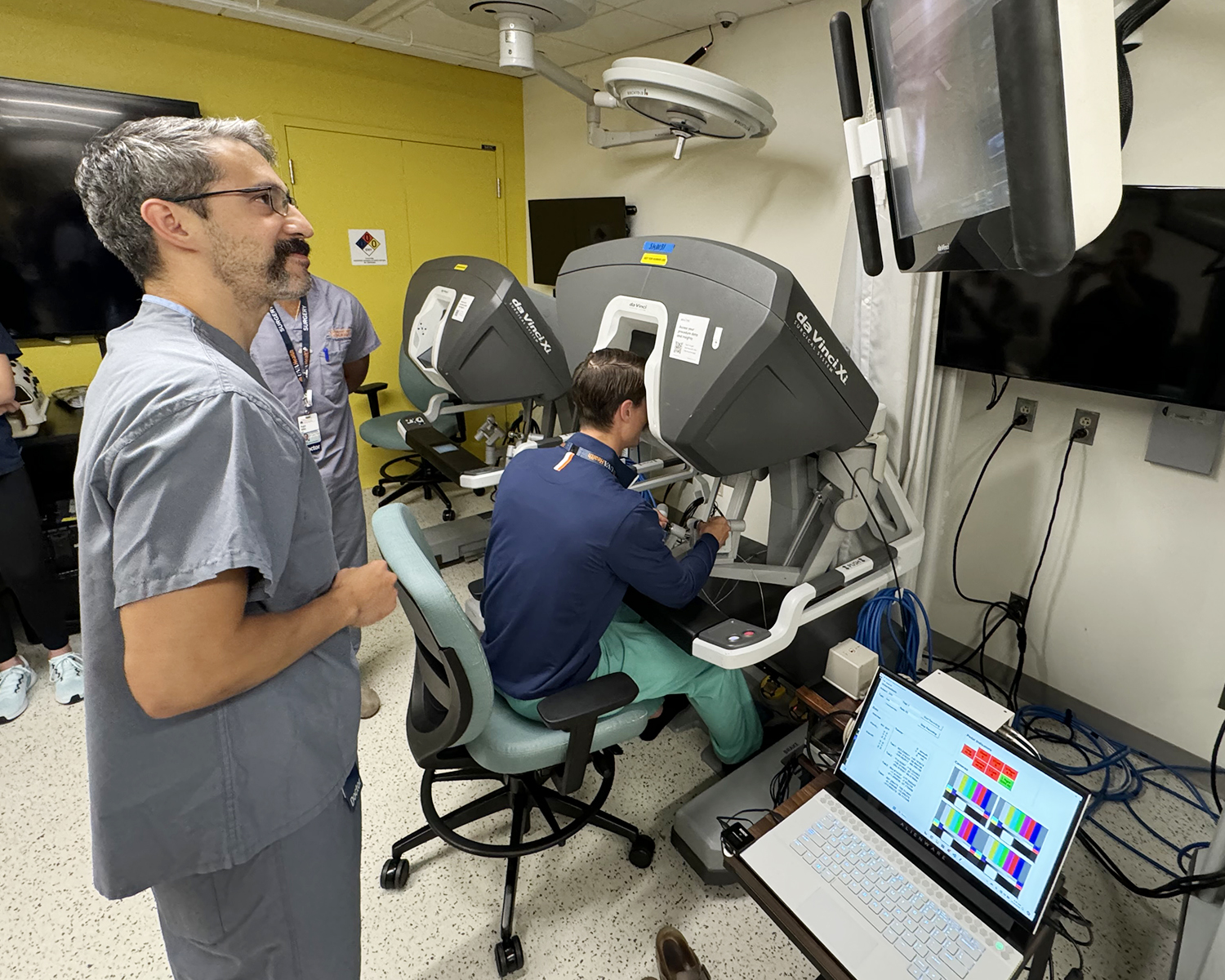}\\[3pt]
        \textbf{(d)} Surgical Bootcamp
    \end{minipage}

    \vspace{5pt}

    \caption{Experimental setups for multimodal data collection on Raven-II and da Vinci Xi systems: (a) Raven-II Peg Transfer task, (b) Raven-II dry-lab setup, (c) da Vinci Xi Suturing task on a realistic hernia repair model, and (d) surgical bootcamp setting for da Vinci Xi data collection.}
    \label{fig:dataset_setup}
\end{figure*}


\subsection{Datasets}

To demonstrate the effectiveness of MiDAS, we collect and publicly release two multimodal datasets acquired on two RAMIS platforms, Raven-II and da~Vinci Xi (see Figure~\ref{fig:dataset_setup}), providing high-fidelity benchmarks for future research. Uniquely, this collection includes synchronized multimodal data recorded during a dry-lab Peg Transfer task and realistic hernia-repair suturing procedures on KindHeart simulation models, bridging the gap between abstract dry-lab tasks and clinically inspired training scenarios. As introduced in Table~\ref{tab:dataset_stats_combined}, the released MiDAS datasets comprise two complementary settings: a Raven-II dataset with internal robot telemetry available for validation, and a da~Vinci Xi dataset collected in a high-fidelity surgical training environment where proprietary robot kinematics were unavailable.

\subsection{Raven-II}

We collected 15 Peg Transfer~\cite{ritter2007design} trials on the Raven-II platform, resulting in 36 minutes of data. We used the DESK taxonomy~\cite{madapana2019desk} for Peg Transfer, which consists of seven gestures covering the sequence of picking up a peg from a starting pole, performing a hand exchange, and placing the peg on a target pole. Each trial involved moving three pegs across the pegboard and was performed by a graduate researcher with prior experience. The peg models and pegboard were selected to resemble those used in practical surgeon training programs. The trials were annotated for gestures by two graduate researchers, with each annotator labeling one half of the data. Ground-truth kinematic data, including MTM and PSM position, orientation, velocity, grasper angle, and clutch pedal presses, were available from the open-source platform and were used for system evaluation.

\subsection{da Vinci Xi} 
\label{daVinci}

We deployed MiDAS during a robotic surgery training bootcamp conducted in 2024 at the University of Virginia (UVA) hospital. Participants performed 7 trials of inguinal hernia repair and 10 trials of ventral hernia repair, among other procedures, using two units of da Vinci Xi systems including surgeon consoles, vision towers and patient-side robotic arms. Throughout the bootcamp, sessions were conducted under direct supervision and guidance of expert resident and attending proctors at UVA, and benefited from high quality KindHeart~\cite{feins2021real} tissue models and standard instruments and endoscopic cameras from Intuitive Surgical. KindHeart provides high-fidelity hernia simulation models using real porcine tissue to closely replicate the anatomical properties and haptic feedback of live surgery. This system ensures realistic tissue handling while maintaining a controlled, repeatable environment essential for consistent data collection.

The resulting data are subsequently annotated and released for public use, aiming to promote the generation of future multimodal surgical datasets and enable development of multimodal cognitive surgical assistants by the research community. The results of a post-completion survey of the participants, and details about the taxonomy of annotations are presented next.

\subsubsection{Participants Feedback Survey}

A total of 40 participants, mostly 2nd and 3rd year surgical residents, performed multiple trials of the surgical tasks described in Section~\ref{daVinci}, 
with a total of $\sim$3.5 hours of annotated suturing segments. 
Survey responses indicated strong acceptance of the simulation environment and supporting infrastructure, with 92.3\% of participants reporting overall satisfaction with the training experience and 86.1\% rating the KindHeart tissue models as realistic. Importantly, \textbf{80\%} of participants \textbf{disagreed or strongly disagreed} that the \textbf{sensors negatively affected their psychomotor performance}, indicating that MiDAS is non-invasive and does not adversely affect participant performance.

\paragraph{Taxonomy of Suturing Gestures}

\begin{figure*}[h]
    \centering
    \includegraphics[width=\textwidth]{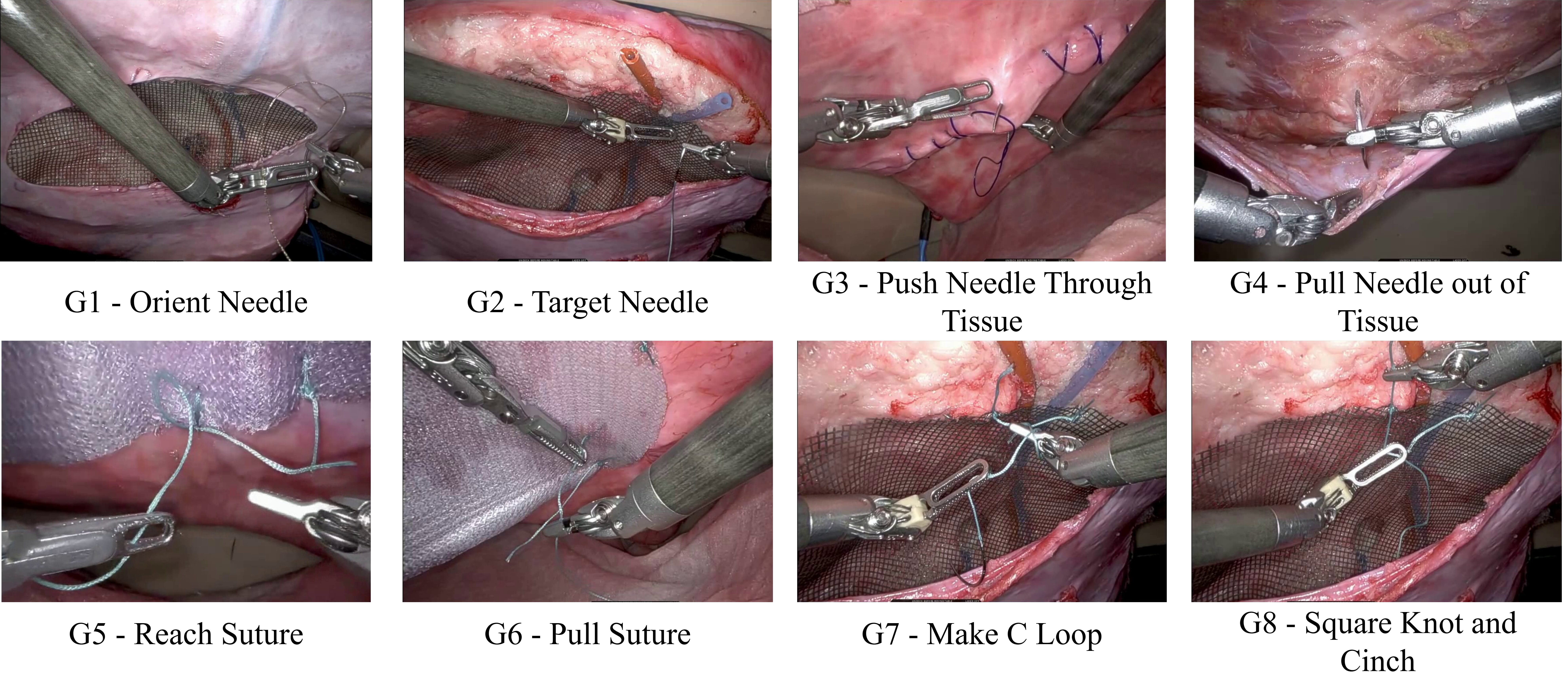}
    \caption{Suturing gestures in Inguinal and Ventral Hernia repair with KindHeart simulation models.}
    \label{fig:gestures}
\end{figure*}

We developed a structured taxonomy of robotic suturing gestures in collaboration with an expert robotic surgeon. The taxonomy builds upon established gesture definitions from prior dry-lab~\cite{gao2014jhu} and in-vivo robotic suturing studies~\cite{psychogyios2023sar}, and was further extended to capture nuances observed in realistic surgical workflows, including fine-grained suture thread management and tissue approximation behaviors (e.g., G6 Pull Suture). The detailed gesture definitions and their distribution within the MiDAS dataset are provided in Table~\ref{tab:surgical_taxonomy} and Table~\ref{tab:dataset_stats_combined}, respectively. Figure~\ref{fig:gestures} shows representative examples of each gesture.

\begin{table*}[t!]
\centering

\begingroup
\footnotesize
\setlength{\tabcolsep}{3pt}
\renewcommand{\arraystretch}{1.08}
\resizebox{\textwidth}{!}{%
\begin{tabular}{@{}p{0.02\textwidth}p{0.15\textwidth}p{0.30\textwidth}p{0.50\textwidth}@{}}
\toprule
\multicolumn{2}{l}{\textbf{Gesture Name}} & \textbf{Goal} & \textbf{Description} \\
\midrule
\textbf{G1} & Orient Needle & Needle is grasped at 50\% distance from tip to end, with a 55$^{\circ}$ angle relative to the grasper. & \textbf{Main arm:} Grasps and manipulates the needle body to adjust its orientation.
\newline \textbf{Support arm:} Grasps the needle temporarily to facilitate re-orientation or hand-off. \\

\midrule

\textbf{G2} & Target Needle & Needle tip contacts the tissue at the intended entry location. & \textbf{Main arm:} Positions the needle tip at the specific entry point on the tissue surface. \newline \textbf{Support arm:} Retracts surrounding tissue to expose the target area or remains idle. \\
\midrule

\textbf{G3} & Push Needle\newline Through\newline Tissue & Needle tip completely penetrates and exits the tissue layers. & \textbf{Main arm:} Applies force to push the needle body through the tissue. \newline \textbf{Support arm:} Retracts tissue to stabilize the entry site or release tension. \\
\midrule

\textbf{G4} & Pull Needle\newline out of Tissue & Needle end fully exits the tissue. & \textbf{Main arm:} Grasps the emerging needle tip and pulls it clear of the tissue. \newline \textbf{Support arm:} Counter-tracts the tissue or assists in grasping the needle if necessary. \\

\midrule

\textbf{G5} & Reach Suture & Suture is grasped at a distance from the tissue that permits loop formation. & \textbf{Main arm:} Maintains position or orients itself to act as the fulcrum for the subsequent loop. \newline \textbf{Support arm:} Grasps an appropriate point on the long end of the suture thread. \\

\midrule

\textbf{G6} & Pull Suture & Suture on the exit side lengthens while the entry side shortens to close the gap. & \textbf{Main arm:} Grasps and pulls the suture thread away from the tissue. \newline \textbf{Support arm:} Either grasps and pulls the suture (in two-handed techniques) or provides a fulcrum point for the thread. \\

\midrule

\textbf{G7} & Make C Loop & A loop of suture is formed around the main instrument. & \textbf{Main arm:} Remains stationary or positions itself to act as the post for the loop formation. \newline \textbf{Support arm:} Manipulates the long end of the suture to wrap it around the opposite jaw. \\

\midrule

\textbf{G8} & Square Knot\newline and Cinch & The knot is formed and tightened securely against the tissue. & \textbf{Main arm:} Grasps the short tail of the suture and pulls it through the loop. \newline \textbf{Support arm:} Pulls the long end of the suture in the opposite direction to cinch the knot. \\

\bottomrule
\end{tabular}%
}
\endgroup
\caption{Taxonomy of surgical gestures for suturing.}
\label{tab:surgical_taxonomy}
\end{table*}

\begin{figure*}[h]
\centering{\includegraphics[width=0.9\linewidth]{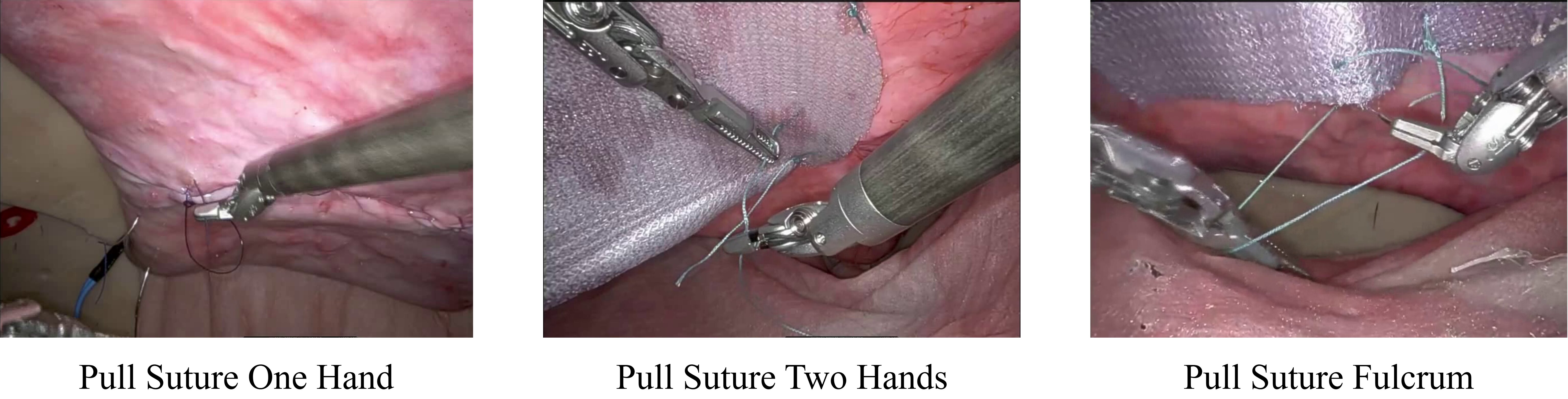}}
\caption{Suture manipulation using one hand, two hands and the fulcrum technique.}
\label{fig:pull_suture}
\end{figure*}

The taxonomy reflects the natural procedural progression of robotic suturing. It begins with needle orientation for proper alignment (G1), followed by needle targeting and positioning at the tissue entry point (G2). The subsequent gestures capture needle driving through tissue (G3) and needle extraction (G4). The workflow then transitions to suture management, including reaching and grasping the suture tail with the opposing instrument (G5), and performing thread manipulation to release slack, maintain workspace organization, or approximate tissue (G6). The final stages include forming the C-loop around the main instrument (G7) and completing square knot tying with cinching (G8). The taxonomy was iteratively refined and validated by a practicing robotic surgeon to ensure clinical relevance.

A distinguishing characteristic of our dataset is the inclusion of intra-gesture variability reflective of real surgical practice. For example, suture manipulation is represented through multiple execution strategies, including single-handed pull, bimanual pull, and fulcrum-assisted pull, as illustrated in Figure~\ref{fig:pull_suture}.

\paragraph{Data Annotation Procedure}
Following our proposed taxonomy, the annotation process was conducted by a multidisciplinary team consisting of an expert surgeon, two graduate researchers, and two undergraduate researchers. The annotators first received training from the expert surgeon, who has extensive experience in laparoscopic and robotic surgery, to ensure consistent interpretation of the gesture definitions and temporal boundaries. To further establish a reference standard for the annotation workflow, the expert surgeon provided complete annotations for two of the largest trials. The remaining videos were divided equally among the annotators, who labeled the raw video data according to the surgeon-guided training and the taxonomy definitions.

An open-source video annotation tool, the VGG Image Annotator (VIA)~\cite{dutta2019vgg}, was customized to align with our taxonomy and to streamline the annotation workflow. Using this platform, annotators marked the temporal gesture boundaries corresponding to when each action began and ended. Each annotation produced by one annotator was reviewed by another annotator, and identified errors were revised. For ambiguous or edge-case segments in which annotators could not confidently identify or segment an action, the expert surgeon provided additional feedback and guidance. These temporally segmented annotations were subsequently processed to generate frame-level gesture labels for downstream analysis.

\begin{figure}[!h]
    \centering
    \includegraphics[width=0.8\columnwidth]{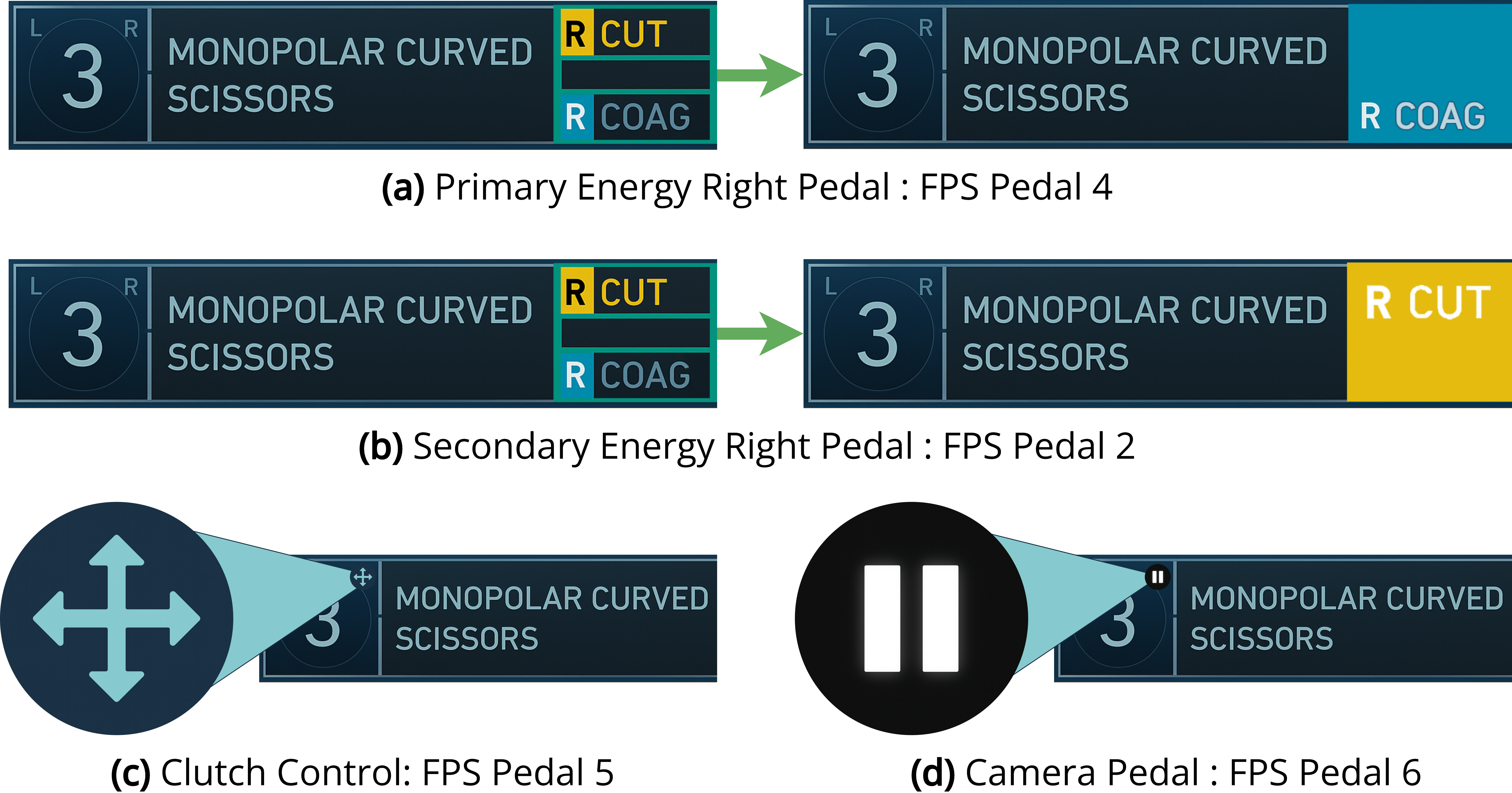}
    \caption{Vision-based extraction of da~Vinci~Xi ground-truth pedal states from surgeon-console UI indicators: (a) Primary Energy Right/FPS Pedal~4 is detected from the right-side \emph{R COAG} activation indicator, (b) Secondary Energy Right/FPS Pedal~2 is detected from the \emph{R CUT} activation indicator, (c) Clutch Control/FPS Pedal~5 and (d) Camera Pedal/FPS Pedal~6 are extracted from their corresponding on-screen icons.}
    \label{fig:pds_gt_vision}
\end{figure}

\paragraph{Pedal Ground-Truth Extraction and Mapping}
The ground-truth internal MTM and PSM kinematic data was not available from da Vinci Xi. We extracted the ground-truth pedal states from the video recordings of surgeon view by analyzing
pedal state indicators (see Figure \ref{fig:pds_gt_vision}). 
Specifically, we extracted pedal states on the user interface by analyzing certain user interface elements similar to~\cite{hashemi2023acquisition}. These elements change color or shape according to the state of their corresponding pedal. We compared the extracted pedal states with the data from the FPS. To match the FPS pedals with those of the da Vinci console, we use the mapping presented in Table \ref{tab:pss_mapping_normalized}.

\begin{table}[!h]
\centering
\caption{Mapping between da~Vinci console pedals and FPS channels with average usage over the Suturing task. Usage is reported both excluding and including the Camera pedal.
{$\blacktriangle$} Not used during Suturing. $\bigstar$ Not detected using FPS, because finger clutch instead of foot pedal clutch was used.}
\label{tab:pss_mapping_normalized}
\begin{tabular}{@{}lcccc@{}}
\toprule
\textbf{da~Vinci pedal} & \textbf{FPS channel} & \textbf{F1} & \multicolumn{2}{c}{\textbf{Pedal Usage (\%)}} \\
\cmidrule(lr){4-5}
 &  &  & \textbf{Energy } & \textbf{All} \\
\midrule
Secondary Energy (L)~$\blacktriangle$    & Pedal~1 & --     & 0.00 & 0.00\\
Secondary Energy (R)  & Pedal~2 & 0.83  & 29.21 & 0.34 \\
Primary Energy (L)~$\blacktriangle$     & Pedal~3 & --     & 0.00 & 0.00 \\
Primary Energy (R)    & Pedal~4 & 0.78  & 70.79 & 4.21 \\
Clutch~$\bigstar$                    & Pedal~5 & --    & --     & 31.03  \\
Camera                    & Pedal~6 & 0.75  & --    & 64.40 \\
Arm Swap~$\blacktriangle$                  & Pedal~7 & --   & --     & 0.00\\
\bottomrule
\end{tabular}
\end{table}



\subsection{Validation and Benchmarking Protocol}\label{experimental_setup}

A key design objective of MiDAS is enabling high-fidelity, non-invasive multimodal data collection without requiring access to proprietary robot telemetry. To establish the reliability of MiDAS as a data acquisition platform, we quantitatively validate externally sensed hand and foot interaction signals against ground-truth internal kinematic and control data available from surgical robotic systems. This validation is critical for demonstrating that externally captured signals can serve as accurate proxies for operator intent and robot execution dynamics.

Specifically, we evaluate the correspondence between MiDAS sensing modalities and internal robot measurements using cross-modal temporal and trajectory alignment analyses. Hand interaction sensing modalities, including EmHT and HandKP, are compared against internal MTM and PSM kinematics. Foot interaction signals captured using FPS are evaluated against ground-truth pedal activation signals recorded from the Raven-II (1 pedal) and da Vinci systems (7 pedals, see Section \ref{daVinci}). Following temporal synchronization, we compute frame-level precision, recall, and F1 score, along with Temporal Intersection over Union (IoU) to measure temporal overlap and residual lag to quantify system delay. The following subsections present detailed correlation, error, and temporal alignment analyses across robotic platforms and tasks.

We benchmark two state-of-the-art (SOTA) temporal gesture recognition models, MTRSAP~\cite{weerasinghe2024multimodal} and MS-TCN++~\cite{farha2019ms}, on MiDAS datasets collected from Peg Transfer (Raven-II) and Suturing (da~Vinci~Xi). Models are trained using (i) single external sensing modalities (EmHT and HandKP), (ii) their fusion with image features (EmHT-Image, HandKP-Image), and are compared against baselines using ground-truth robot kinematics (MTM and PSM on Raven-II) and image-only inputs. Proprietary MTM/PSM kinematics were unavailable for da~Vinci~Xi; therefore, image features serve as the only baseline for that platform. Raven-II data were downsampled from 30~Hz to 10~Hz, while da~Vinci~Xi data were used at their native 30~Hz. All models were developed and evaluated using identical architectures, training protocols, and train/validation/test splits to ensure fair comparison.

We adopt a gesture-level train/validation/test split for gesture recognition experiments. For all gesture recognition experiments,   All trials are first segmented into individual gesture instances, which are then partitioned in a stratified manner by gesture class. Specifically, we follow a 60/20/20 split. This stratified splitting preserves class balance across all sets and evaluates performance on held-out gesture instances rather than held-out trials. 

To further assess baseline validity and the effect of dataset size, we additionally evaluate kinematic- and image-only models trained on larger related datasets: DESK~\cite{madapana2019desk} (from COMPASS~\cite{hutchinson2023compass}) for Peg Transfer, and SAR-RARP50~\cite{psychogyios2023sar} and SAIS~\cite{Kiyasseh2023SAIS} for Suturing, used solely for contextual comparison with prior work.

We also performed a comprehensive ablation study to evaluate the effect of different kinematic feature sets, including velocity, grasper angle, and clutch states, as well as different visual representation backbones. Specifically, we compared two widely adopted visual encoders: Residual Network-50 (ResNet-50)~\cite{wang2018deep}, a CNN architecture commonly used for image feature extraction, and DINOv2~\cite{oquab2023dinov2}, a self-supervised vision transformer model that learns general-purpose visual representations. We further evaluated the effect of domain-adaptive pretraining using larger surgical activity datasets, such as DESK~\cite{madapana2019desk}. More details are presented in \textit{Appendix B}.


\section{Results}

The Results section is structured to correspond directly to the evaluation protocols detailed in Section~\ref{experimental_setup} of the Materials and Methods. Section~\ref{emht_handkp_correlation_analysis} presents quantitative validation of externally sensed hand trajectories against internal robot kinematics, followed by Section~\ref{pedal_data}, which reports performance metrics for pedal state detection. Section~\ref{gesture-recognition-results} presents gesture recognition benchmark results using the MiDAS datasets, including modality comparisons and ablation analyses.

\subsection{Correlation Analysis}


\subsubsection{Electromagnetic and Depth-Based Hand Tracking}
\label{emht_handkp_correlation_analysis}

\begin{figure*}[!t]
    \centering
    \includegraphics[width=\textwidth]{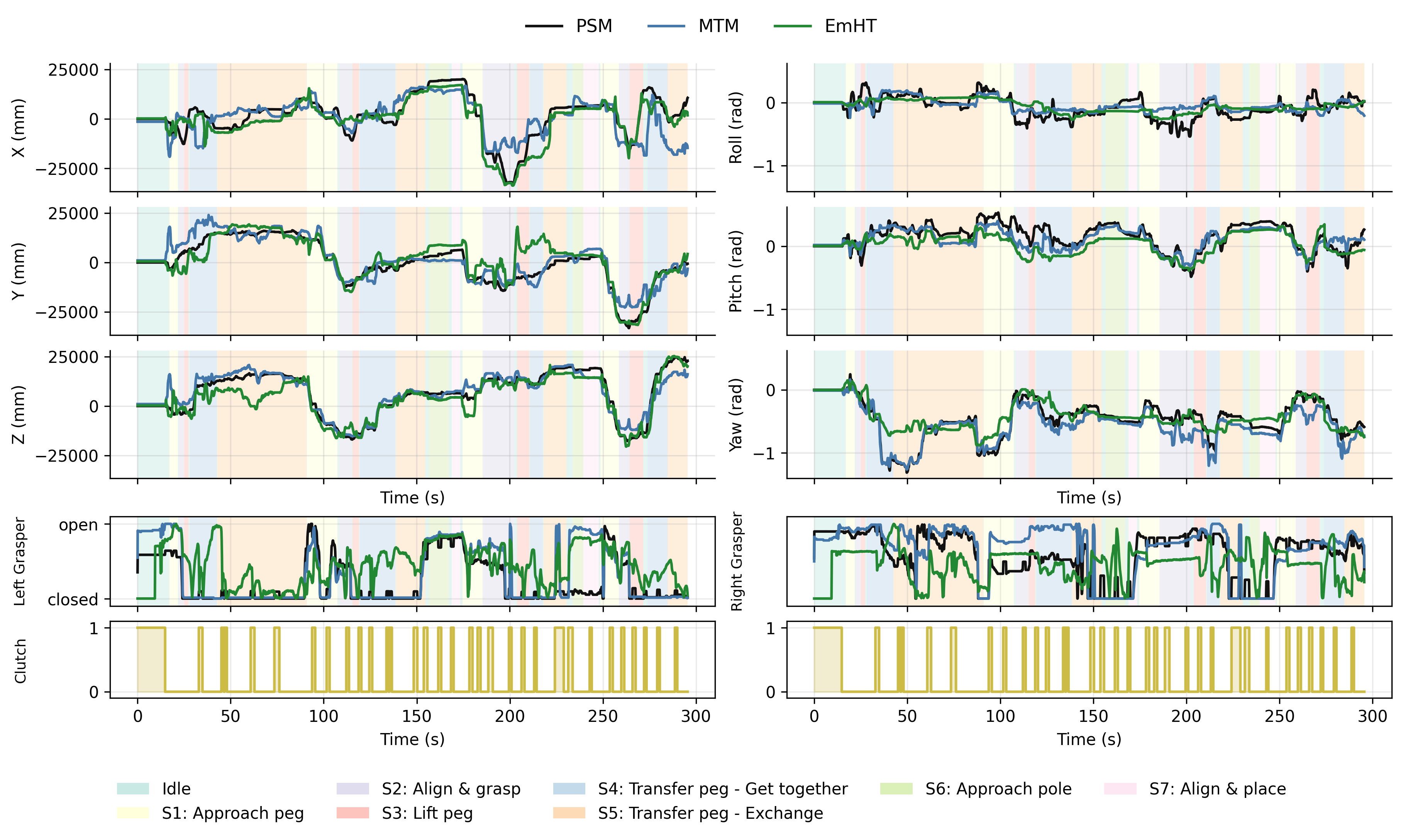}
    \caption{MTM and EmHT alignment with PSM for one robotic arm across position, orientation, grasper, and clutch signals during a Peg Transfer trial.}
    \label{fig:corr_visualization}
\end{figure*}

\begin{figure}[t!]  
\centering
\includegraphics[width=0.8\columnwidth]{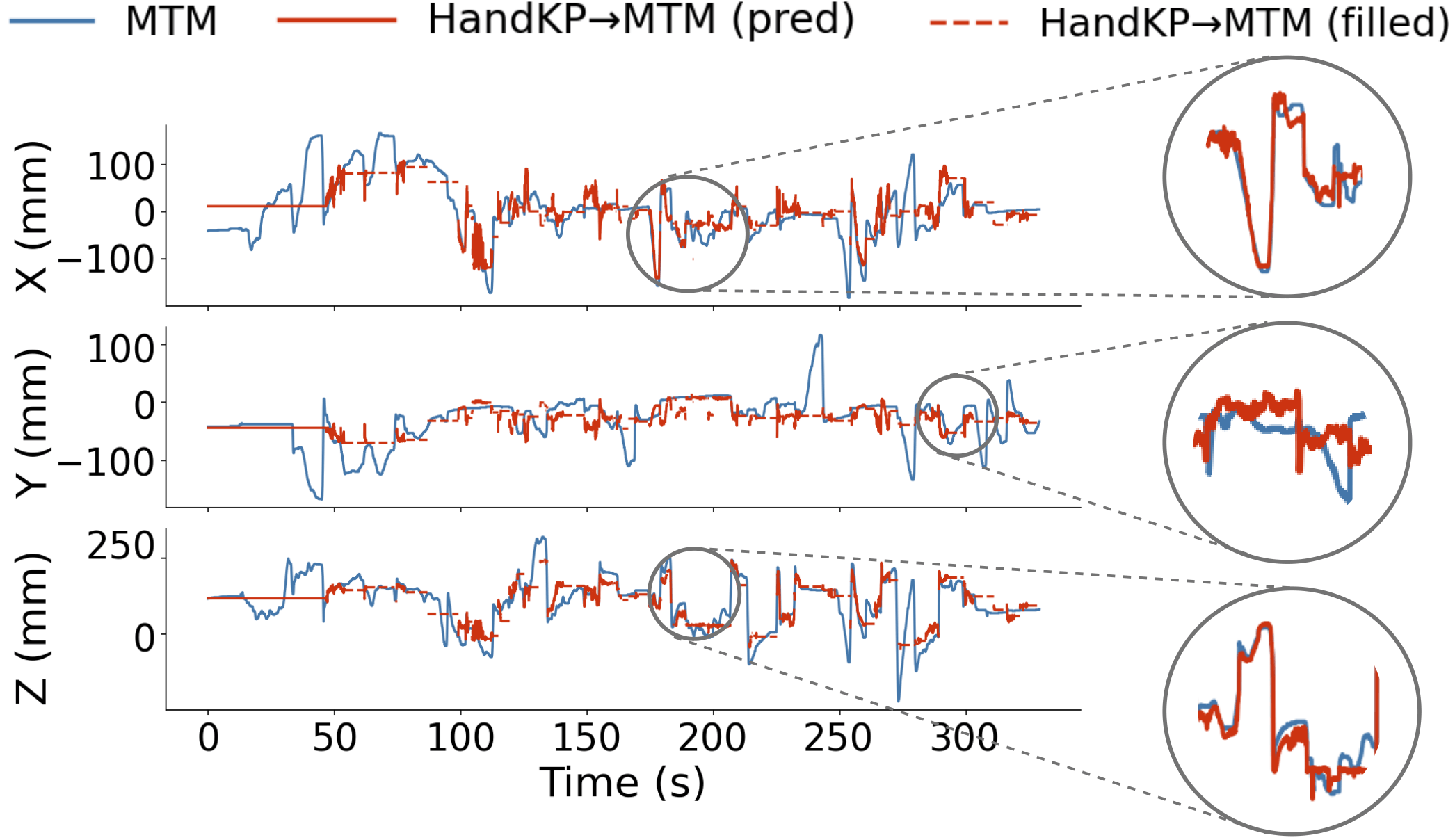}
\caption{Estimated left hand keypoint trajectory (Left HandKP) vs. Left MTM. Average of thumb and middle finger keypoints for each hand are transformed to the MTM frame. The gaps caused by missed hand keypoint detection are filled using interpolation for short gaps and forward filling for longer gaps.}
\label{fig:hand_traj}
\end{figure}

As summarized in Table~\ref{tab:corr_metrics}, both EmHT and HandKP show strong positional correspondence with ground-truth MTM and PSM. EmHT captures the translational motion patterns of both the operator's input (MTM) and the final tool-tip (PSM) (Figure~\ref{fig:corr_visualization}), with \textbf{EmHT-MTM} and \textbf{EmHT-PSM} mean CoS exceeding \textbf{0.8 across all axes}. HandKP also achieves robust positional similarity to MTM and PSM trajectories (see Figure~\ref{fig:hand_traj} for MTM alignment), with \textbf{HandKP-MTM} and \textbf{HandKP-PSM} mean CoS \textbf{exceeding 0.7} for all axes and \textbf{0.88 Z-axis alignment}. A more in-depth analysis is provided in \textit{Appendix A.3}.

\begin{table}[t]
\caption{Mean CoS and NRMSE(\%) of EmHT--MTM, EmHT--PSM, HandKP--MTM, and HandKP--PSM for positional ($X$, $Y$, $Z$) and rotational (Roll, Pitch, Yaw) trajectories over 15 Peg Transfer trials. Arrows indicate whether higher ($\uparrow$) or lower ($\downarrow$) values are desirable.}
\label{tab:corr_metrics}
\centering
\begin{tabular}{@{}lcccccc@{}}
\toprule
\multirow{2}{*}{\textbf{Modalities}} & \multicolumn{2}{c}{\textbf{X}} & \multicolumn{2}{c}{\textbf{Y}} & \multicolumn{2}{c}{\textbf{Z}} \\
\cmidrule(lr){2-7}
  & CoS$\uparrow$ & NRMSE$\downarrow$ & CoS$\uparrow$ & NRMSE$\downarrow$ & CoS$\uparrow$ & NRMSE$\downarrow$ \\
\midrule
EmHT-MTM & \textbf{0.81} & \textbf{17.5} & \textbf{0.86} & \textbf{24.4} & \textbf{0.88} & \textbf{16.2} \\
EmHT-PSM & \textbf{0.82} & \textbf{17.0} & \textbf{0.80} & \textbf{21.0} & \textbf{0.89} & \textbf{16.4} \\
\cmidrule(lr){2-7}
HandKP-MTM & 0.72 & 26.13 & 0.71 & 30.15 & \textbf{0.89} & 19.88 \\
HandKP-PSM & \textbf{0.74} & \textbf{24.03} & \textbf{0.72} & \textbf{26.71} & 0.88 & \textbf{19.21} \\
\midrule
\midrule
\multirow{2}{*}{\textbf{Modalities}} & \multicolumn{2}{c}{\textbf{Roll}} & \multicolumn{2}{c}{\textbf{Pitch}} & \multicolumn{2}{c}{\textbf{Yaw}} \\
\cmidrule(lr){2-7}
  & CoS$\uparrow$ & NRMSE$\downarrow$ & CoS$\uparrow$ & NRMSE$\downarrow$ & CoS$\uparrow$ & NRMSE$\downarrow$ \\
\midrule
EmHT-MTM & \textbf{0.65} & 50.6 & \textbf{0.61} & \textbf{23.5} & \textbf{0.90} & \textbf{24.8} \\
EmHT-PSM & 0.32 & \textbf{44.1} & 0.44 & 38.4 & 0.36 & 36.9 \\
\bottomrule
\end{tabular}%
\end{table}

For orientation, EmHT captures MTM trends with reasonable fidelity: \textbf{EmHT-MTM} achieves a \textbf{CoS of 0.90 for Yaw}, while Pitch and Roll show moderate alignment (0.61 and 0.65, respectively). In contrast, \textbf{EmHT-PSM} orientation alignment is lower, with CoS values ranging from 0.32 (Roll) to 0.44 (Pitch). Orientation estimation from HandKP data did not achieve strong correlations with ground-truth; corresponding results are reported in \textit{Appendix A.3}.

For grasper angle, EmHT achieves moderate agreement with both MTM and PSM grasper angles (IoU $\approx$ 0.53--0.54, Accuracy $\approx$ 0.80--0.81). Additional details on grasper-angle estimation from EmHT are provided in \textit{Appendix A.2.2}.

\subsubsection{Foot Pedal State Detection}
\label{pedal_data}

\begin{figure*}[!t]
    \centering

    \begin{minipage}{0.48\linewidth}
        \centering
        \includegraphics[width=\linewidth]{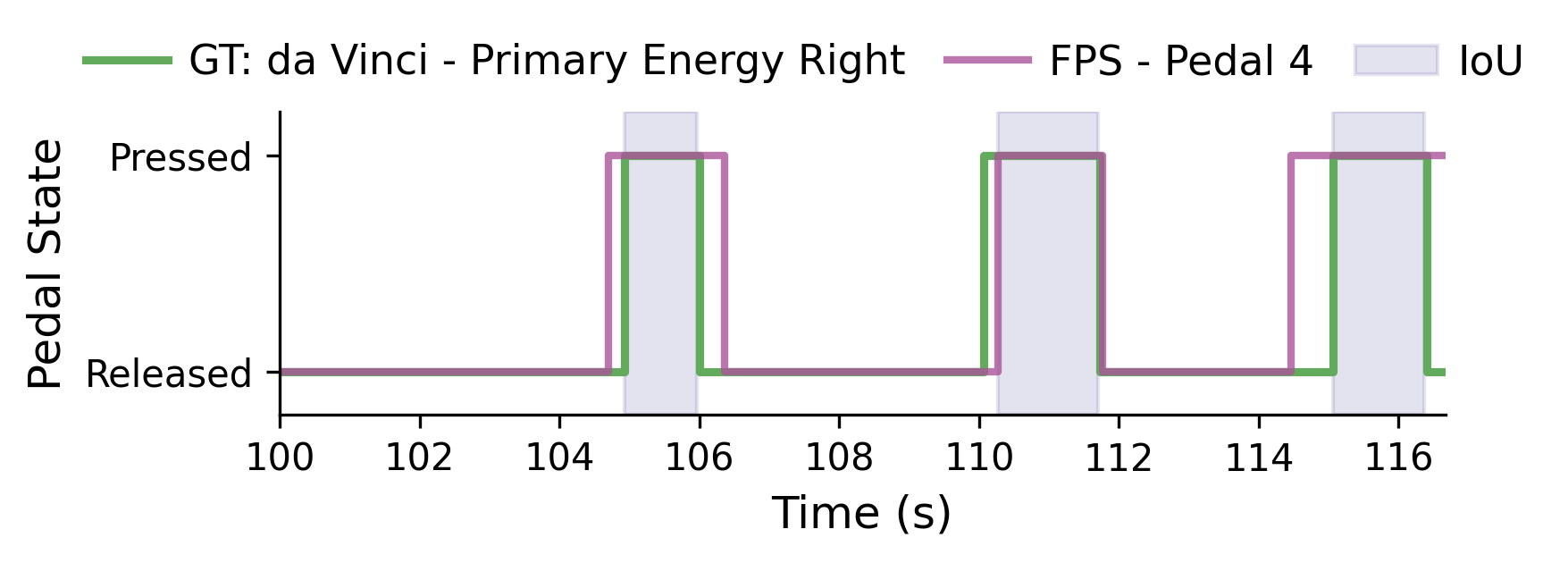}
        \textbf{(a)} da Vinci Xi Primary Energy Right GT vs PSS detection
    \end{minipage}
    \hfill
    \begin{minipage}{0.48\linewidth}
        \centering
        \includegraphics[width=\linewidth]{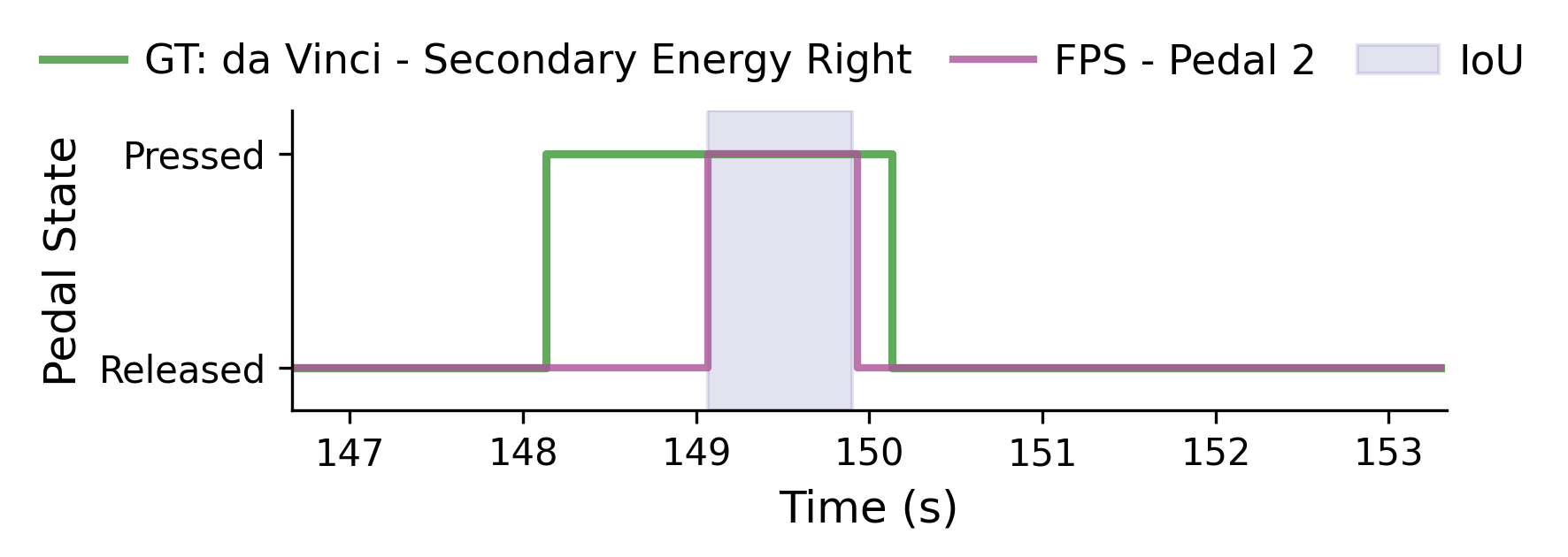}
        \textbf{(b)} da Vinci Xi Secondary Energy Right GT vs PSS detection
    \end{minipage}

    \vspace{10pt}

    \begin{minipage}{0.48\linewidth}
        \centering
        \includegraphics[width=\linewidth]{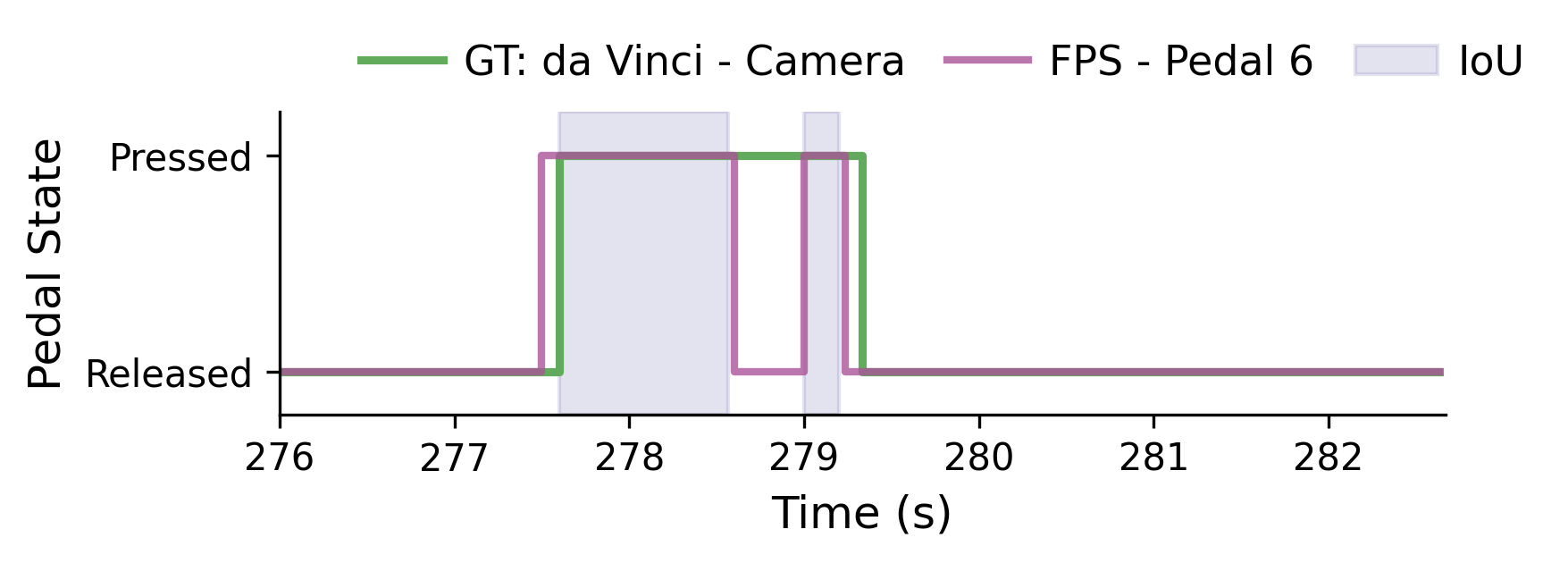}
        \textbf{(c)} da Vinci Xi Camera Pedal GT vs PSS detection
    \end{minipage}
    \hfill
    \begin{minipage}{0.48\linewidth}
        \centering
        \includegraphics[width=\linewidth]{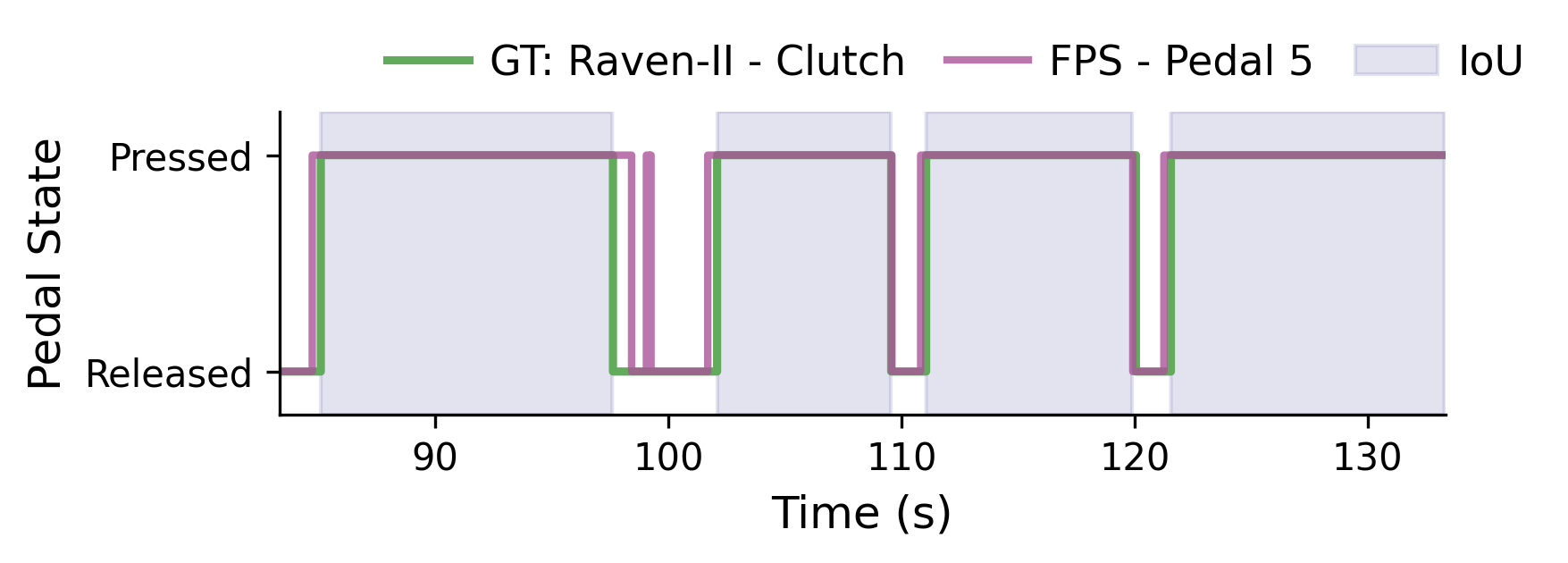}
        \textbf{(d)} Raven-II Clutch Pedal GT vs PSS detection
    \end{minipage}
    \vspace{5pt}

    \caption{Representative foot pedal state detection against ground-truth (GT), with shaded regions indicating temporal IoU overlap. On da~Vinci Xi, (a) Primary Energy Right/Pedal~4 shows close alignment across repeated activations, (b) Secondary Energy Right/Pedal~2 captures the main activation with a small onset delay, and (c) Camera/Pedal~6 illustrates strong overlap with a brief transition-level mismatch. On Raven-II, (d) Clutch/Pedal~5 shows comparable alignment over longer activation intervals. Together, the panels demonstrate strong temporal alignment during active periods, with occasional false positives or missed samples near state transitions.}   
    \label{fig:pds_vis}
\end{figure*}

Figure~\ref{fig:pds_vis} presents representative step plots of ground-truth and predicted pedal states for both systems. On da~Vinci Xi, panels (a)-(c) show that FPS captures primary-energy, secondary-energy, and camera-pedal activations with strong temporal alignment, while small onset/offset mismatches illustrate occasional transition-level errors. Panel (d) shows comparable alignment for the Raven-II clutch signal during longer activation intervals, supporting cross-platform pedal state sensing.

\begin{table}[!h]
\caption{Foot pedal state detection performance. Metrics are averaged across multiple trials.}
\label{tab:pds_results}
\vspace{1em}
\centering
\begin{tabular}{@{}lccccc@{}}
\toprule
\textbf{Platform} & \textbf{F1} & \textbf{Precision} & \textbf{Recall} & \textbf{IoU} & \textbf{Lag (ms)} \\
\midrule
Raven-II    & 0.85 & 0.78 & 0.95 & 0.74 & 166.67 \\
da~Vinci Xi & 0.78 & 0.69 & 0.86 & 0.67 & 133.33 \\
\bottomrule
\end{tabular}%
\end{table}

Table~\ref{tab:pds_results} summarizes the quantitative results aggregated across multiple trials. 
For Raven-II, FPS achieves high recall (0.95), indicating that pedal presses are rarely missed, while maintaining moderate precision (0.78), reflecting some false positives. This results in a strong F1 score of 0.85, demonstrating robust overall detection performance despite significant class imbalance favoring the unpressed state. The Temporal IoU of 0.74 indicates that approximately 74\% of active pedal time is correctly captured by the system. The measured lag of 166.7\,ms represents the end-to-end detection latency, which includes sensor sampling, processing, and synchronization delays.

For da Vinci Xi, the system maintains high recall (\(0.86\)) with moderate precision (\(0.69\)), reflecting a bias toward capturing all true pedal activations at the expense of occasional false positives. This balance yields an F1 score (\(0.78\)), indicating strong overall detection performance across pedals.

We also evaluated the per-pedal usage and detection accuracy for da Vinci Xi experiments (see Table~\ref{tab:pss_mapping_normalized}). Pedal usage was measured using ground-truth data extracted from video and varied by procedure and console configuration. For example, \emph{Camera} pedal was most frequently used, but the \emph{Clutch} pedal was infrequently actuated because clutching was available on the hand manipulators. In suturing trials, considering energy pedals only, \emph{Primary Energy (Right)} mapped to FPS \emph{Pedal~4} accounted for roughly \(70.8\%\) of all actuations with a detection F1 of 0.78.



\subsection{Downstream Evaluation for Gesture Recognition}
\label{gesture-recognition-results}

We evaluated the performance of non-invasive sensing modalities (EmHT and HandKP) against ground-truth robot kinematics and baseline vision-only approaches. Table \ref{tab:abl_modality} summarizes the results for the Peg Transfer task on the Raven-II platform using two temporal backbone models, MTRSAP~\cite{weerasinghe2024multimodal} and MS-TCN++~\cite{farha2019ms}.

As expected, models trained on internal robot telemetry (MTM and PSM) established a high baseline, with MTRSAP achieving an F1-score of 0.87 for both MTM and PSM inputs. The non-invasive EmHT modality demonstrated performance closely matching these proprietary signals, achieving an accuracy of 0.87 and an F1-score of 0.86 with the MTRSAP model. In contrast, the HandKP modality resulted in significantly lower performance (F1-score 0.38), and image-only models (using DINOv2 features) achieved an F1-score of 0.48 due to the limited dataset size (15 trials).

Table \ref{tab:abl_modality_suturing_louo} presents the results for the Suturing task on the da Vinci Xi. Since internal robot kinematics were unavailable for this clinical system, we compared EmHT against image-based baselines. The fusion of EmHT with visual features (\textbf{EmHT, Image}) yielded the highest performance across both models, with MTRSAP achieving an accuracy of 0.71 and F1-score of 0.70. This outperformed the single-modality Image baseline (Accuracy 0.69) and the EmHT-only baseline (Accuracy 0.64).

\begin{figure}[!t]
    \centering
    \includegraphics[width=0.8\columnwidth]{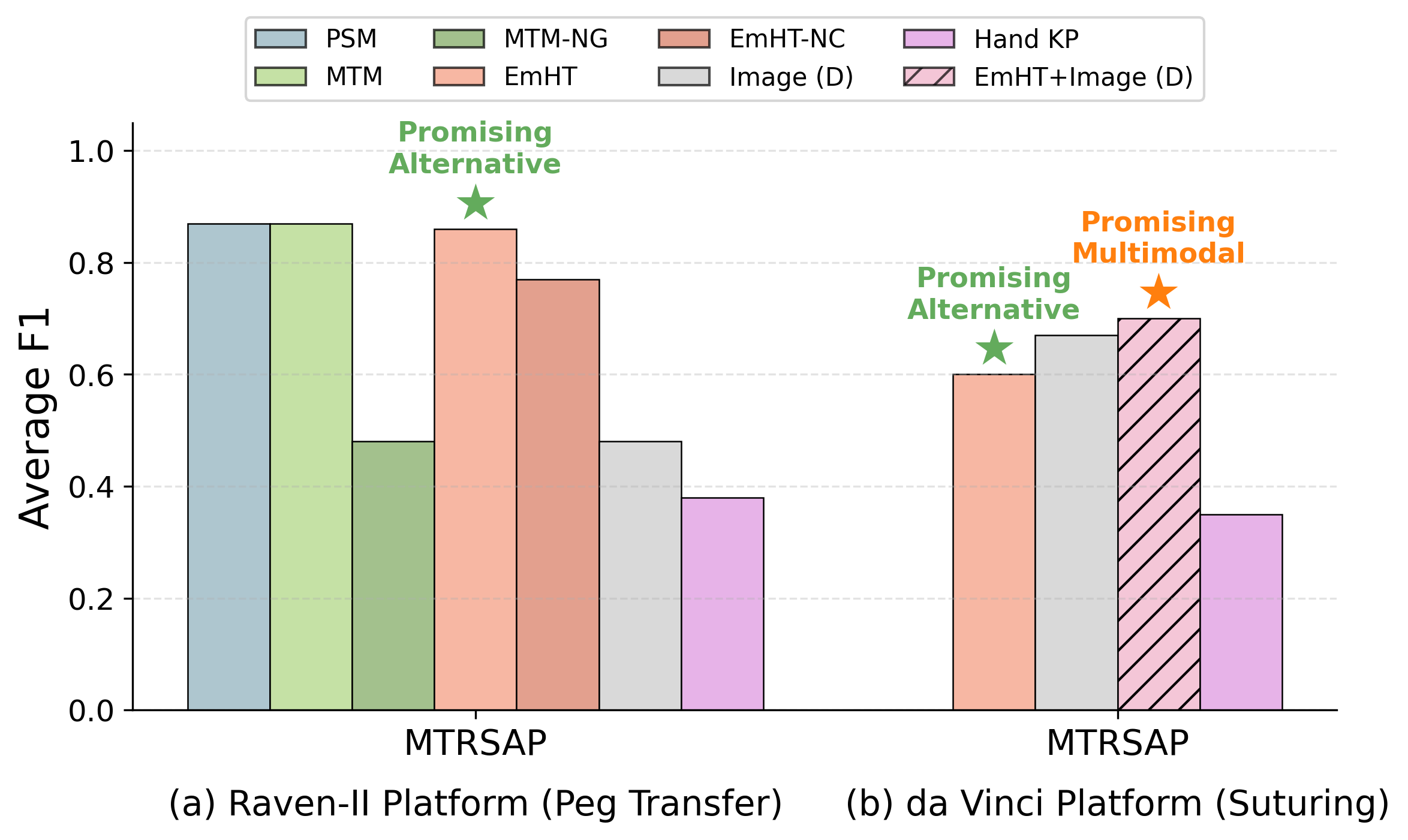}
\caption{Best average F1 scores across selected modalities on both platforms. (a) Raven-II Peg Transfer includes internal PSM/MTM baselines; EmHT is nearly competitive with proprietary kinematics, supporting non-invasive motion sensing as a kinematic proxy. (b) da~Vinci Xi Suturing has no PSM/MTM comparison because proprietary robot kinematics were unavailable during data collection; EmHT+Image achieves the highest F1, showing that external motion cues complement visual features. \textit{Abbreviations: NG--No Grasper Angle; NC--No Clutch State; (D)--DINOv2.}}
    \label{fig:model_accuracy_comparison}
\end{figure}

Figure~\ref{fig:model_accuracy_comparison} summarizes the best average F1 scores across selected modalities. In panel (a), Raven-II Peg Transfer results show that EmHT performs nearly on par with internal PSM/MTM kinematics, while image-only and HandKP baselines achieve lower performance. In panel (b), da~Vinci Xi Suturing results show that EmHT+Image achieves the strongest performance in the absence of internal robot telemetry, highlighting the complementary value of external motion sensing and visual features.

\begin{table*}[!t]
\centering
\caption{Peg Transfer on Raven-II gesture recognition using different modalities. Best performing modality/feature setting is reported. \textbf{Bold} values indicate the best-performing feature configuration for selected modalities within each model. \(\dagger\) Image-only modalities underperform due to limited data. * Same baselines evaluated on a larger dataset for comparison. \textit{Abbreviations: pos.=position; vel.=velocity; ori.=orientation; grasper.=grasper angle.}}
\label{tab:abl_modality}
\resizebox{1\textwidth}{!}{%
\begin{tabular}{@{}llllcccc@{}}
\toprule
\textbf{Dataset} & \textbf{Model} & \textbf{Modality} & \textbf{Features} &
\textbf{Acc} & \textbf{F1} & \textbf{Prec.} & \textbf{Rec.} \\
\midrule

\multirow{14}{*}{\centering\arraybackslash\shortstack[l]{\textbf{Raven-II}\\[1pt]Peg Transfer\\[0pt](15 trials)}}
 & \multirow{8}{*}{\textbf{MTRSAP}~\cite{weerasinghe2024multimodal}}

 & \textbf{PSM}        & \textbf{pos.,  ori., vel., grasper.} 
                &\textbf{0.88}  & \textbf{0.87} & \textbf{0.88}  & \textbf{0.87} \\
\cmidrule(l){3-8}

& & \textbf{MTM}        & \textbf{pos., ori., vel., grasper.} 
                &  0.87 & 0.87  & 0.90  & 0.86  \\
\cmidrule(l){3-8}

 & & Image$\dagger$ & RGB (DINOv2)
                & 0.48 & 0.48 & 0.48 & 0.48 \\
 \cmidrule(l){3-8}

 & & \textbf{EmHT} & \textbf{pos., ori., vel.} 
                & \textbf{0.87} & \textbf{0.86} & \textbf{0.87} & \textbf{0.86} \\
                
 & & EmHT,Image & pos., ori., vel., RGB (DINOv2)
                & 0.58 & 0.56  &  0.60 &  0.58 \\
 
\cmidrule(l){3-8}

 & & HandKP     & pos., vel. 
                & 0.40 & 0.38 & 0.41 & 0.39 \\

  & & HandKP, Image    & pos., vel., RGB (DINOv2)
                &  0.38 & 0.33  & 0.36  & 0.37  \\

\cmidrule(l){2-8}

 & \multirow{6}{*}{\textbf{MS-TCN++}~\cite{farha2019ms}}
   & \textbf{PSM}        & \textbf{pos., ori., vel., grasper.} 
                & 0.85 & 0.84 & 0.86 & 0.86 \\
\cmidrule(l){3-8}

 & & \textbf{MTM} & \textbf{pos., ori., vel., grasper.} 
                & \textbf{0.86} & \textbf{0.86} & \textbf{0.86} & \textbf{0.87} \\

\cmidrule(l){3-8}

 & & Image$\dagger$ & RGB (DINOv2)
                & 0.40 & 0.47 & 0.41 & 0.44 \\
                
\cmidrule(l){3-8}

 & & \textbf{EmHT} & \textbf{pos., ori., vel.} 
                & \textbf{0.79} & \textbf{0.78} & \textbf{0.80} & \textbf{0.81} \\

 & & EmHT,Image & pos., ori., vel., RGB (DINOv2)
                & 0.65  & 0.64 &  0.67 & 0.67   \\
 
\cmidrule(l){3-8}

 & & HandKP     & pos., vel.
                & 0.33 & 0.29 & 0.27 & 0.37 \\
                
  & & HandKP, Image  & pos., vel., RGB (DINOv2)
                         & 0.49  & 0.51  & 0.49  &  0.46 \\            

\midrule

\multirow{4}{*}[-0.5ex]{\centering\arraybackslash\shortstack[l]{\textbf{DESK~\cite{madapana2019desk}}\\[1pt]Peg Transfer\\[0pt](48 trials)}}
 & \multirow{2}{*}{\textbf{*MTRSAP}~\cite{weerasinghe2024multimodal}}
   & PSM        & pos., vel., ori., grasper. 
                & 0.91 & 0.91 & 0.92 & 0.92 \\
                \cmidrule(l){3-8}
 & & Image  & RGB (DINOv2) 
                & 0.78 & 0.78 & 0.77 & 0.78 \\
\cmidrule(l){2-8}
 & \multirow{2}{*}{\textbf{*MS-TCN++}~\cite{farha2019ms}}
   & PSM        & pos., vel., ori., grasper. 
                & 0.92 & 0.93 & 0.92 & 0.92 \\
                \cmidrule(l){3-8}
 & & Image  & RGB (DINOv2) 
                & 0.74 & 0.78 & 0.76 & 0.77 \\            
\bottomrule
\end{tabular}%
}
\end{table*}

\begin{table*}[!t]
\centering
 \caption{Suturing gesture recognition on the MiDAS da Vinci Xi dataset across different modality and feature configurations. For each model on the MiDAS da Vinci Xi dataset, the best-performing configuration by accuracy is highlighted in \textbf{bold}.
 Results on SAR-RARP50 and SAIS are included only as larger baseline datasets for comparison. 
\textit{Abbreviations: pos.=position; vel.=velocity; ori.=orientation; $\dagger$=As reported or estimated from published results.}}
\label{tab:abl_modality_suturing_louo}
\resizebox{1\textwidth}{!}{%
\begin{tabular}{@{}llllcccc@{}}
\toprule
\textbf{Dataset} & \textbf{Model} & \textbf{Modality} & \textbf{Features} &
\textbf{Acc} & \textbf{F1} & \textbf{Prec.} & \textbf{Rec.} \\
\midrule

\multirow{15}{*}{\centering\arraybackslash\shortstack[l]{\textbf{da Vinci Xi}\\[1pt]Suturing\\[0pt](17 trials)}}
 & \multirow{8}{*}{\textbf{MTRSAP}~\cite{weerasinghe2024multimodal}}
   
  & Image           & RGB (DINOv2)
                         & 0.69 & 0.67 & 0.65 & 0.65 \\
\cmidrule(l){3-8}

  & & EmHT                & pos., ori., vel.
                         & 0.64 & 0.60 & 0.63 & 0.61 \\

  & & \textbf{EmHT, Image} 
                         & \textbf{pos., ori., vel., RGB} 
                         & \textbf{0.71} & \textbf{0.70} & \textbf{0.70} & \textbf{0.69} \\
\cmidrule(l){3-8}

       & & HandKP     & pos., vel.
                 & 0.40 & 0.35 & 0.37 & 0.36 \\
                         
 & & HandKP, Image & pos., vel., RGB
                         & 0.58 & 0.58 & 0.58 & 0.57 \\

\cmidrule(l){2-8}
 & \multirow{7}{*}{\textbf{MS-TCN++}~\cite{farha2019ms}}

  & Image           & RGB (DINOv2)
                         & 0.57 & 0.57 & 0.53 & 0.51 \\

\cmidrule(l){3-8}

   &  & EmHT                & pos., ori., vel. 
                         & 0.57 & 0.50 & 0.50 & 0.53 \\

 & & \textbf{EmHT, Image} 
                         & \textbf{pos., ori., vel., RGB}
                         & \textbf{0.63} & \textbf{0.56} & \textbf{0.64} & \textbf{0.63} \\
\cmidrule(l){3-8}
                         
    & & HandKP     & pos., vel.
                 &  0.43 & 0.36  & 0.39  & 0.37  \\

  & & HandKP, Image  & pos., vel., RGB 
                         & 0.58 & 0.54  &  0.58  &  0.54  \\

\midrule

\multirow{2}{*}{\centering\arraybackslash\shortstack[l]{\textbf{SAR-RARP50~\cite{psychogyios2023sar}}\\[0pt]Suturing (50 trials)}} 
 & \textbf{ASFormer~\cite{yi2021asformer}} & Image 
    & RGB {(I3D~\cite{carreira2017quo})}
    & $0.82^{\dagger}$ & -- & -- & -- \\
 & \textbf{MViT~\cite{fan2021multiscale}} & Image 
    & RGB {(ViT\cite{dosovitskiy2020image})}
    & $0.79^{\dagger}$ & -- & -- & -- \\
\midrule

\multirow{2}{*}[0.2ex]{\centering\arraybackslash\shortstack[l]{\textbf{SAIS~\cite{Kiyasseh2023SAIS}}\\[0pt]Suturing (78 trials)}} 
 & \multirow{2}{*}{\shortstack[l]{\textbf{Vision}\\\textbf{Transformer}~\cite{Kiyasseh2023SAIS}}}
   & Image & RGB 
   & -- & $0.50^{\dagger}$ & -- & -- \\
 & & \multicolumn{6}{l}{} \\
\bottomrule
\end{tabular}%
}
\end{table*}

\section{Discussion}

\noindent\textbf{Correlation Analysis.}
The correlation analysis results demonstrate that externally sensed motion signals collected by MiDAS closely approximate internal robot telemetry in both trajectory alignment and temporal correspondence.
EmHT achieved high cosine similarity and low normalized RMSE when compared against internal MTM/PSM kinematics, indicating strong agreement in translational motion patterns. While orientation-related signals and grasper angle estimation exhibited comparatively higher error, the overall alignment trends remained consistent across tasks. In contrast, RGB-D keypoint–based HandKP measurements showed reduced robustness, primarily due to missed detections and occlusions when hands moved outside the camera field of view. These findings confirm that external EmHT provides a stable and reliable surrogate for internal kinematic measurements, particularly for translational components of surgeon motion.

Similarly, validation results of the FPS support its reliability as a non-invasive proxy for console pedal states. Frame-level precision, recall, and F1 scores indicate strong agreement with ground-truth pedal activations across both Raven-II and da Vinci systems, while temporal Intersection over Union (IoU) and residual lag analyses show minimal delay and accurate event boundary alignment. Minor discrepancies can be attributed to brief transitional states and synchronization offsets, but these do not materially affect overall activation detection performance. Together, the correlation analysis for hand motion and the high classification fidelity of pedal states demonstrate that MiDAS can capture surgeon interaction signals with sufficient accuracy to approximate proprietary telemetry, supporting its use in downstream learning and analysis tasks without direct access to internal robot data.

\noindent\textbf{Implications for Downstream Tasks.}
The gesture recognition experiments further validate the practical utility of the proposed non-invasive modalities. On the Raven-II platform, EmHT achieved gesture recognition performance comparable to that of internal robot kinematics (MTM and PSM), with only a marginal reduction in F1-score (0.86 vs. 0.87). This finding is significant as it suggests that low-dimensional kinematic proxies derived from EmHT are sufficiently rich to model complex surgical dynamics, even without direct access to the robot's internal state.

Conversely, the RGB-D based HandKP modality performed suboptimally (F1-score of 0.38). This drop in performance correlates with the occlusion issues identified in the trajectory analysis, confirming that line-of-sight constraints severely limit the reliability of optical hand tracking for continuous activity recognition in this setting.

In the clinical context of the da Vinci Xi, where proprietary kinematic data was unavailable, the fusion of EmHT with visual features provided the most robust performance (F1-score 0.70), surpassing both vision-only and EmHT-only baselines. This aligns with prior literature suggesting that multimodal fusion is particularly beneficial in low-data regimes where individual modalities may be noisy or incomplete. The results demonstrate that MiDAS provides a viable pathway for conducting high-fidelity multimodal research on clinical systems, offering robust non-invasive alternatives to restricted proprietary telemetry.

\noindent\textbf{Limitations and Future Work.}
The datasets collected in this study are moderate in scale, consisting of 15 Peg Transfer trials on the Raven-II platform and 17 Suturing trials on the da Vinci Xi system collected during a controlled surgical training bootcamp. While sufficient to validate the feasibility of the MiDAS system and provide initial benchmarks for multimodal gesture recognition, they remain limited in size compared with large-scale surgical video and computer vision datasets. In addition, the data were collected in controlled training environments with a limited number of participants, which may not fully capture the variability present across surgeons, procedures, and real clinical workflows.

The scale of the current datasets also has implications for model evaluation and generalizability. In this study, gesture recognition experiments use stratified train/validation/test splits over gesture instances to preserve class balance and enable initial comparisons across sensing modalities. However, given the limited number of trials and participants, this evaluation setting may not fully measure generalization across unseen trials, users, or institutions. Larger future MiDAS datasets would enable more stringent validation protocols, such as leave-one-trial-out~\cite{rivas2024instrument}, leave-one-supertrial-out, and leave-one-user-out~\cite{gao2014jhu,hutchinson2023compass} evaluation, which would provide stronger evidence of model robustness in unseen settings. Larger datasets would also support more reliable analysis of rare gestures, inter-participant variability, and procedure-specific differences.

Another direction for future extensions of MiDAS is the integration of additional sensing modalities. The current system focuses on non-invasive motion, pedal interaction, and surgical video streams, which are directly related to gesture recognition and robot-assisted task execution. However, physiological and behavioral metrics associated with surgeon workload, stress~\cite{takacs2024assessment}, and fatigue, such as heart rate, heart-rate variability, respiration, eye gaze, and blink behavior, could provide complementary context for interpreting surgical performance during training. Incorporating these signals in future MiDAS-based datasets may support more comprehensive analysis of how operator state relates to gesture execution, skill evaluation, error patterns, and workflow progression in robot-assisted surgery.

The dataset-scale limitations and the expansion of sensing modalities  discussed above are closely tied to the practical challenges of multimodal data collection in surgical settings, including restricted access to robotic platforms, the cost and logistical complexity of wet-lab experiments, and the limited availability of surgical residents for extended data collection sessions. Importantly, the goal of this work is not to introduce a large-scale dataset, but rather to present a generalizable, non-invasive data acquisition framework and to demonstrate its feasibility through representative multimodal datasets. Building on the current benchmarks, future work will focus on scaling data collection across more participants, procedures, and clinical environments, enabling more rigorous validation while extending MiDAS-based datasets and benchmarks to support broader applications such as objective surgical skill assessment, error detection, and workflow analysis.

\noindent\textbf{Conclusion.}
This work presents \textbf{MiDAS}, a platform-agnostic, non-invasive multimodal data acquisition system for robot-assisted surgery. MiDAS enables synchronized collection of EmHT, optical RGB-D keypoints, and pedal interaction signals without access to proprietary robot hardware or software. Correlation analysis against Raven-II ground-truth kinematics shows that EmHT reliably approximates robot end-effector motion, achieving sub-20\% normalized RMSE and high cosine similarity. Downstream gesture recognition experiments further demonstrate that external sensing modalities, particularly EmHT, achieve performance \emph{competitive} with proprietary robot kinematics and outperform vision-based approaches due to their robustness to occlusions and out-of-view conditions. These results highlight non-invasive motion sensing as a practical surrogate for internal robot telemetry in real surgical environments. By releasing the MiDAS system, multimodal datasets, and baseline benchmarks, we aim to support reproducible, cross-platform research, reduce reliance on closed robotic systems, and accelerate progress in multimodal learning for surgical activity understanding.


\subsection*{Author Contributions}
The authors confirm contribution to the paper as follows: 
System conception and design: Keshara Weerasinghe, Seyed Hamid Reza Roodabeh, Zachary Schrader, Andrew Hawkins, Homa Alemzadeh.
Data collection:  Keshara Weerasinghe, Seyed Hamid Reza Roodabeh, Andrew Hawkins, Zachary Schrader, Homa Alemzadeh.
Analysis and interpretation of results: Keshara Weerasinghe, Seyed Hamid Reza Roodabeh, Andrew Hawkins, Zhaomeng Zhang, Homa Alemzadeh.
Manuscript preparation:  Keshara Weerasinghe, Seyed Hamid Reza Roodabeh, Zhaomeng Zhang, Andrew Hawkins, Homa Alemzadeh.
Supervision and project administration: Homa Alemzadeh.
All authors reviewed the results and approved the final version of the manuscript.

\subsection*{Funding \& Acknowledgements}
This work was partially supported by research awards from National Science Foundation (CNS-2146295) and the Commonwealth Cyber Initiative. We thank Kay Hutchinson for her contributions to the early conceptualization of MiDAS, exploration of initial sensing approaches, and early drafts of this work. We also thank Shrisha Yapalparvi and Sara Liu for their contributions to early system developments and experiments; Tessa Heick for assistance with data collection; and Alex Farmer and Maximilian Meer for data annotation. Special thanks to Christopher Scott, MD (now at Mayo Clinic Jacksonville), Shane Strike and Mike Lewis (Intuitive Surgical), and the University of Virginia Department of Surgery for their help with organizing the Bootcamp.

\subsection*{Ethical Statement}
The surgical bootcamp and associated data collection were approved by the University of Virginia Institutional Review Board under protocol IRB-SBS 6724. All participants provided informed consent prior to participation.

\subsection*{Conflict of Interest}
The authors declare that they have no conflict of interest.

\subsection*{Data Availability}
The multimodal datasets and data collection system introduced in this study are publicly available at \url{https://uva-dsa.github.io/MiDAS/}.

\subsection*{Patient Consent}
Not applicable. The study involved surgical residents performing training tasks on simulation models and did not involve patient data.

\subsection*{Permission to Reproduce Material}
The authors confirm that all figures, tables, and materials are original or reproduced with appropriate permissions.

\subsection*{Clinical Trial Registration}
Not applicable. This study does not involve a registered clinical trial.

\clearpage


\startcontents[after]           
\setcounter{tocdepth}{2}        

\clearpage
\appendixformat

\begin{center}
    \LARGE \textbf{Technical Appendix} \\[1em]
    \large MiDAS: A Multimodal Data Acquisition System and Dataset\\ for Robot-Assisted Minimally Invasive Surgery
\end{center}
\vspace{1em}

\printcontents[after]{}{1}{\section*{Table of Contents}}

\section{Data Collection System \& Correlation Analysis}
\label{dcs}


\subsection{High-level Architecture}

\begin{figure}[htbp]
\begin{center}
\centerline{\includegraphics[width=0.7\columnwidth]{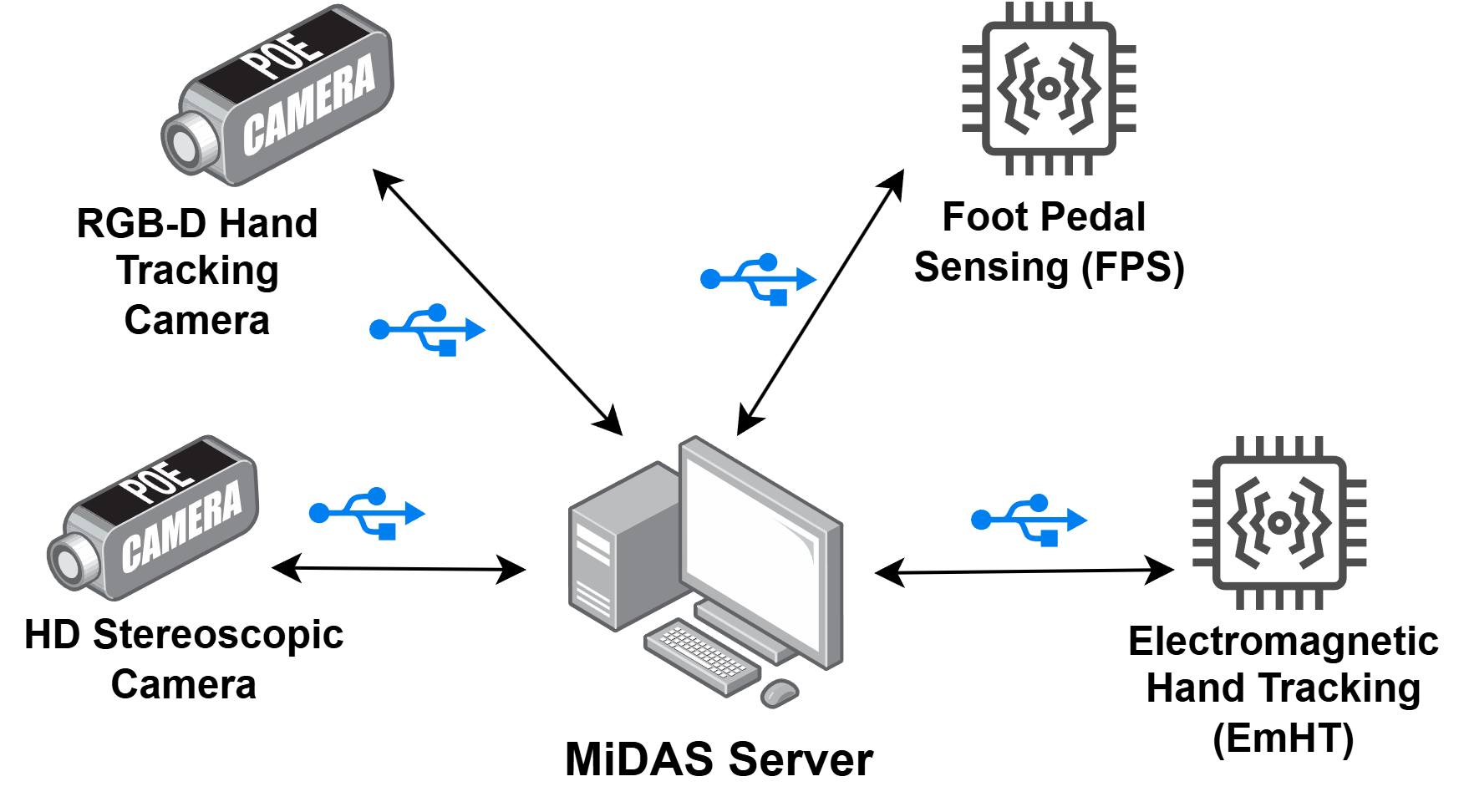}}
\caption{High-level architecture of the MiDAS, illustrating the client-server design with multiple sensing modalities, including an HD stereoscopic camera, an RGB-D hand tracking camera, Electromagnetic Hand Tracking (EmHT) sensors, and Foot Pedal Sensing (FPS) system, all streaming data to the centralized MiDAS server.}
\label{fig:midas-arch}
\end{center}
\end{figure}

\begin{figure}[h!]
\begin{center}
\centerline{\includegraphics[width=1\columnwidth]{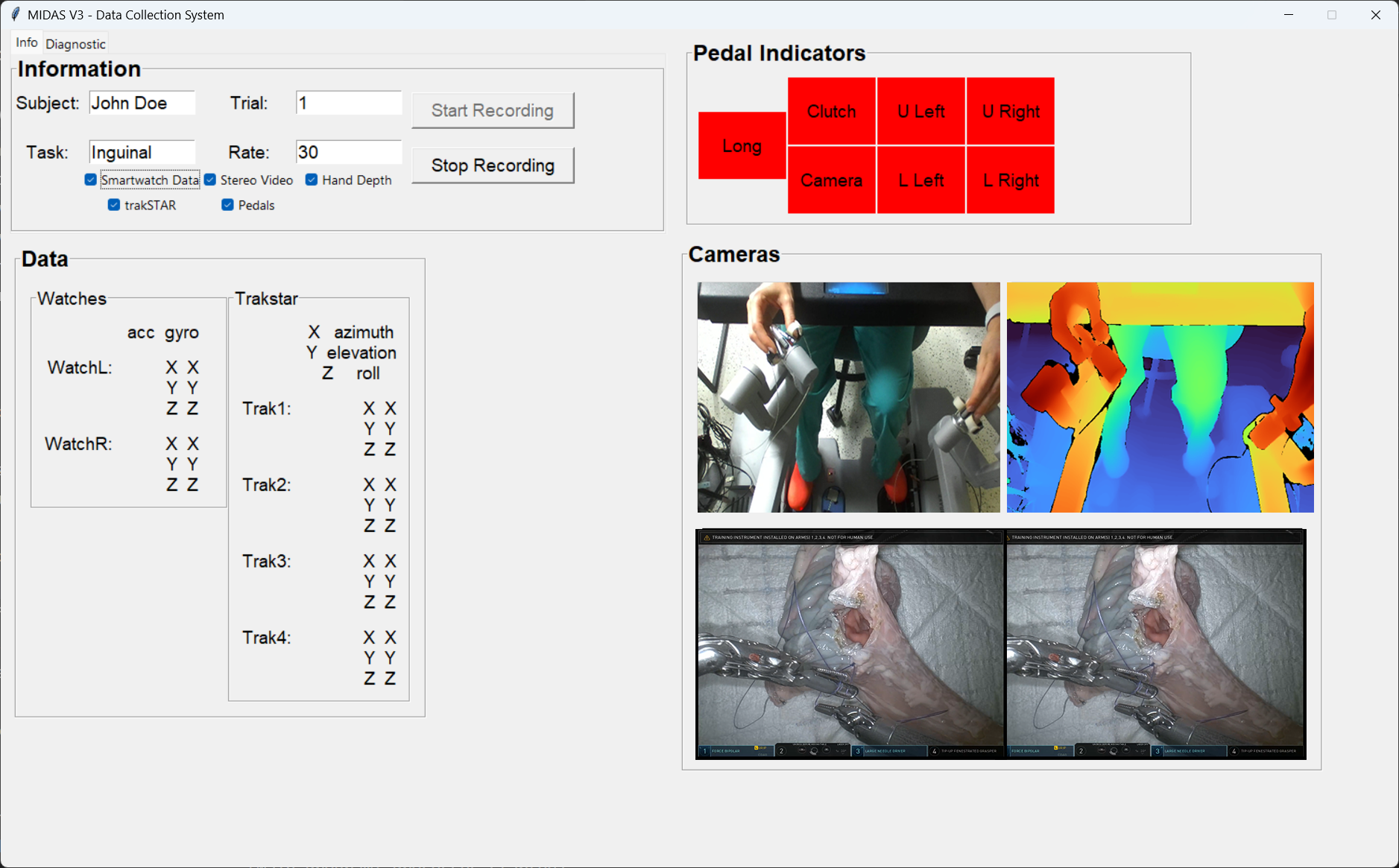}}
\caption{GUI of MiDAS Server. The procedure information such as subject, trial, task and acquisition rate are entered by the user. The various pedal states, values of EmHT sensors, RGB-D view of hands and stereo surgical view are displayed in real-time.}
\label{fig:midas-gui}
\end{center}
\end{figure}

We adopt a client-server architecture for data acquisition. A central server coordinates multiple client processes that stream heterogeneous data (e.g., kinematics, video) as shown in Figure \ref{fig:midas-arch} at different rates, implemented via Python \texttt{multiprocessing} workers to enable concurrent, real-time ingestion and buffering. The server provides a GUI (Figure \ref{fig:midas-gui}) to (i) enter study metadata (subject, procedure, trial) and master sampling frequency \(f\); (ii) select modalities to be acquired at the target rate \(f\); and (iii) monitor live system status of all connected clients through a health dashboard. The modular code, implemented in Python, separates transport, scheduling, and I/O components, allowing easy extension to new modalities and deployment scenarios.

\subsubsection{Data Synchronization}
All streams are stamped on arrival at the MiDAS server with a Unix epoch timestamp, providing a unified, server-side time base that is insensitive to client clock drift. During post-processing, we treat the OBS video frame timestamps as the reference clock and align all other modalities to this time base via their server timestamps (causal frame-wise association to the nearest video time). We then visualize the synchronized signals (video, kinematics, and pedal sensor traces) for manual spot-checks to confirm alignment quality at frame granularity.

\subsubsection{Cost Estimation}
A key design goal of MiDAS was to minimize system cost in order to encourage broad adoption by the research community. At the time of development and experimentation (Q1 2024), the estimated cost for a single surgeon-console MiDAS setup was approximately \$8,000 (USD). This cost includes an NDI \emph{trakSTAR} 3D Guidance electronics unit and accessories ($\sim$\$4,850), Blackmagic SDI recorders and cables for stereoscopic video capture (2$\times$, $\sim$\$400 total), a ZED Mini RGB-D camera for hand-depth capture ($\sim$\$400), a custom foot pedal sensing system ($\sim$\$150), and an appropriate laptop workstation to host the MiDAS server ($\sim$\$2,000). Overall, the complete hardware setup totals approximately \$8,000 at the time of development.

\subsubsection{Data Recording Performance}
When recording at a master synchronization frequency of 30 Hz, MiDAS captures stereoscopic surgical scene video at 1080p resolution, Electromagnetic Hand Tracking (EmHT) data, RGB-D hand depth data, and foot pedal sensing signals, resulting in an aggregate storage requirement of approximately 0.4GB/minute. This data rate can be reduced by lowering the recording frequency and/or decreasing the resolution of the surgical scene video, depending on the requirements of the target application.

\subsubsection{Data Recording Latency}
Data recording latency is quantified as the average time delay between sensor data transmission from the acquisition device to the MiDAS server and the subsequent persistent storage operation. For example, electromagnetic hand tracking data acquired by the \emph{trakSTAR} 3D Guidance electronics unit are transmitted to the MiDAS server via USB serial communication. The acquisition unit assigns a transmission timestamp to each data packet, and the MiDAS server appends a secondary timestamp upon committing the data to disk.

This source–destination timestamp–based mechanism is employed for all sensing modalities where device-side (source) timestamps are available to facilitate latency characterization. For modalities that do not expose source-side timestamps, such as surgical scene video captured via SDI output from the da Vinci vision cart and recorded through OBS, only destination side timestamps at write time are available, and thus comparable end-to-end recording latency cannot be reliably estimated under this definition.

Studies indicate surgeons have average baseline reaction times (318–330 ms) that slow significantly under cognitive load~\cite{pfister2014comparison}. During active robotic manipulation, dual-task experiments show reaction times extending to ~0.4–0.9 s, increasing with complexity~\cite{lim2023physiological, yang_reaction}. Thus, the practical operative response window (400–900 ms) is substantially slower than the isolated ~300 ms baseline. Based on the mean recording latencies for each sensing modality reported in Table~\ref{tab:latencies} and processing times that are within few milliseconds scale, we propose the feasibility of MiDAS for real-time downstream applications, including real-time online gesture recognition, real-time error detection, and other time-critical processing tasks.

Note that reported latencies may depend on the hardware configuration of the MiDAS host system. All measurements in this study were obtained using an Alienware m15 R7 laptop equipped with a 12th Generation Intel\textsuperscript{\textregistered} Core\texttrademark{} i7-12700H processor, NVIDIA\textsuperscript{\textregistered} GeForce\texttrademark{} RTX 3070 Ti GPU (8~GB GDDR6), 32~GB DDR5 memory, and a 2~TB PCIe Gen4 NVMe solid-state drive.

\begin{table*}[]
    \centering
    \caption{Average data recording latency across MiDAS sensing modalities. Latency is computed using source-to-disk timestamps for modalities that expose device-side timestamps. $^{\dagger}$Surgical scene video captured via SDI and recorded through OBS does not expose source-side timestamps, and therefore comparable end-to-end recording latency cannot be reliably estimated under the timestamp-based definition.}
    \label{tab:latencies}
    \begin{tabular}{llc}
        \toprule
        \textbf{Modality} & \textbf{Sensing Device} & \textbf{Average Latency (ms)} \\
        \midrule
        Electromagnetic Hand Tracking & NDI \emph{trakSTAR}~\cite{NDI3DGuidance} & 43.3 \\
        
        Hand Depth & ZED Mini Stereo Camera~\cite{StereolabsZEDMini} & 24.66 \\
        
        Surgical Scene Video$^{\dagger}$ & Blackmagic SDI Recorder (OBS) & N/A \\
        
        Foot Pedal States & Foot Pedal Sensing (FPS) & 12.97 \\
        \bottomrule
    \end{tabular}

\end{table*}



\subsection{3-Dimensional Electromagnetic Hand Tracking}
\label{sec:emht}
This section provides implementation details for the EmHT module described in Section~\ref{sec:emht}. The NDI \emph{trakSTAR} electromagnetic tracking device~\cite{NDI3DGuidance} uses four miniature 6-DoF sensors mounted on the MTM controls, with two sensors per hand. Sensors are indexed as: (1) left middle finger, (2) left thumb, (3) right thumb, and (4) right middle finger. This placement reflects typical console usage, with thumb/middle-finger grip and index-finger clutching, and enables continuous finger-level pose traces during operation.

According to the manufacturer, the nominal translation range of the system is $\pm 76$~cm in any direction, with an angular range of $\pm 180^\circ$ in azimuth and roll and $\pm 90^\circ$ in elevation under ideal conditions. When the field generator is mounted near the MTM console at the resting height of the joysticks, this range is sufficient to encompass the typical console hand workspace. However, the effective tracking volume and accuracy depend on local electromagnetic distortion from nearby metallic and electronic components, so the final placement should be validated experimentally in the target operating room layout. The trakSTAR system is controlled via the vendor C++ API and integrated into the MiDAS acquisition pipeline.

\subsubsection{Electromagnetic Hand Tracking for Pose Estimation}
MiDAS logs the 6-DoF pose for each sensor ($i=1...4$) at the 270 Hz sampling rate, expressed in the EmHT's coordinate frame ($T$). This pose, $\mathbf{P}_{T,i}$, combines a 3D position vector $\mathbf{p}_{T,i}$ and a 3D orientation vector $\boldsymbol{\theta}_{T,i}$:
\begin{align}
    \mathbf{p}_{T,i} &= [x_i, y_i, z_i]^T \in \mathbb{R}^3 \\
    \boldsymbol{\theta}_{T,i} &= [\alpha_i, \epsilon_i, \rho_i]^T \in \mathbb{R}^3
\end{align}
where $(\alpha_i, \epsilon_i, \rho_i)$ represent azimuth, elevation, and roll, respectively.

These externally sensed poses can be mapped to the coordinate frames of the Master Tool Manipulator (MTM), denoted $M$, and the Patient Side Manipulator (PSM), denoted $P$. This mapping is primarily achieved via a rigid body transformation, $\mathbf{T}$, which is composed of a rotation matrix $\mathbf{R}$ and a translation vector $\mathbf{t}$. This transformation can be obtained or estimated for any robot through a calibration procedure. For Raven-II robot, we obtain the transformations from the open source MTM and control software code bases. To compensate for modeling uncertainties and sensor/controller inaccuracies, we augment this with a learned residual term, $f(\mathbf{p}; \boldsymbol{\phi})$, modeled as a simple Multi-Layer Perceptron (MLP) network, with $\boldsymbol{\phi}$ representing the learnable parameters of the model.

For example, the position of a sensor $\mathbf{p}_{T,i}$ can be transformed into the MTM coordinate frame ($M$) to find the corresponding position $\mathbf{p}_{M,i}$ using the rigid transformation and a residual model $f_M$ with parameters $\boldsymbol{\phi}_M$:
\begin{equation}
    \mathbf{p}_{M,i} = \mathbf{R}^{M}_{T} \, \mathbf{p}_{T,i} + \mathbf{t}^{M}_{T} + f_M(\mathbf{p}_{T,i}; \boldsymbol{\phi}_M)
\end{equation}
Similarly, a transformation maps the sensor's position into the PSM coordinate frame ($P$) using $\mathbf{T}^{P}_{T}$ (composed of $\mathbf{R}^{P}_{T}$ and $\mathbf{t}^{P}_{T}$) and a residual $f_P$ with parameters $\boldsymbol{\phi}_P$:
\begin{equation}
    \mathbf{p}_{P,i} = \mathbf{R}^{P}_{T} \, \mathbf{p}_{T,i} + \mathbf{t}^{P}_{T} + f_P(\mathbf{p}_{T,i}; \boldsymbol{\phi}_P)
\end{equation}
Both MLP models $f_P$ and $f_M$ have 2 hidden layers with 16 neurons and use Rectified Linear Units as activation function. A similar composition of rotations is used to transform the orientation $\boldsymbol{\theta}_{T,i}$ into the target frames. The models take a pose in EmHT coordinates, and transform it onto the MTM or PSM coordinates. The training objective function is Root Mean Square Error (RMSE) with $L_2$ weight decay and 50\% dropout to prevent over-fitting.
Training is performed using leave-one-trial-out cross-validation (LOTO-CV) over the 15 Raven-II Peg Transfer trials. In each fold, one complete trial is held out for validation and the model is trained on the remaining 14 trials. Metrics are computed on the held-out trial and averaged across all 15 folds. A summary of model and training hyperparameters are presented in Table \ref{tab:mlp_params}.

\begin{table}[htbp]
\centering
\caption{Hyperparameters and training configuration for the residual MLP model.}
\label{tab:mlp_params}
\vspace{1em}
\begin{tabular}{@{}ll@{}}
\toprule
\textbf{Parameter} & \textbf{Value} \\
\midrule
Number of Hidden Layers & 2 \\
Neurons per Layer & 16 \\
Activation Function & ReLU \\
Dropout Probability & 0.5 \\
\midrule
Objective Function & RMSE \\
Regularization & $L_2$ Weight Decay \\
Weight Decay Coeff. & 1.5 \\
Optimizer & Adam \\
Learning Rate & 0.005 \\
Epochs & 50 \\
\bottomrule
\end{tabular}
\end{table}

\subsubsection{Electromagnetic Hand Tracking for Grasper Angle Estimation}
\begin{figure}[htbp]
    \centering
    \includegraphics[width=\textwidth]{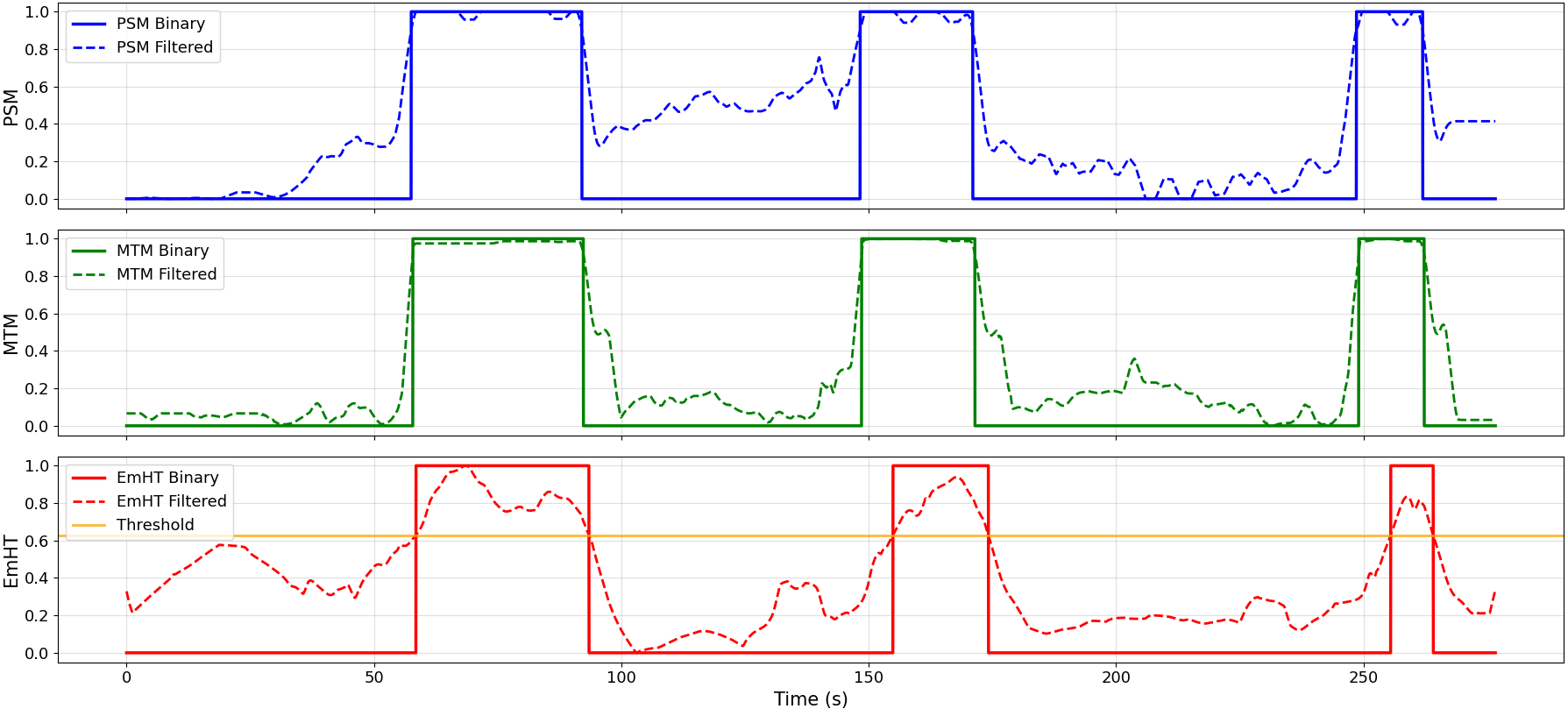}
    \caption{Left grasper angle comparison among the PSM, MTM, and EmHT measurements. Solid lines represent binarized open (0) /closed (1) states; dashed lines show normalized filtered raw signals; orange line marks the EmHT closure threshold.}
    \label{fig:gripper_status_analysis}
\end{figure}

In addition to the trajectory analysis, we also conducted an analysis to assess whether the EmHT sensors mounted on the MTM can reliably estimate the actual grasper open/close states of both the MTM and the PSM. To determine whether each grasper was open or closed, we first computed the Euclidean distance between the two EmHT sensors on the MTM grasper over time. A Moving Average filter was then applied to smooth the raw signal and highlight the underlying trend. Lastly, an appropriate threshold was selected to convert the filtered EmHT signal into a binarized open/close state, as illustrated in Figure~\ref{fig:gripper_status_analysis}.

According to Table \ref{tab:gripper_analysis}, the MTM and PSM grasper angles showed strong consistency (IoU: 0.90, Accuracy: 0.97, Precision: 0.97), which means that the PSM grasper accurately executes the MTM commands. However, the EmHT sensors achieved moderate agreement with both MTM and PSM grasper angles (IoU $\approx$ 0.53--0.54, Accuracy $\approx$ 0.80--0.81), indicating that while EmHT can generally capture overall open/close behavior, it is not a direct replacement for capturing precise grasper angles due to signal noise and sensor placement strategy.

\begin{table*}[htbp]
\caption{Mean Intersection over Union (IoU), Accuracy, and Precision metrics for grasper angle estimation across the PSM–MTM, EmHT–MTM, and EmHT–PSM modalities, computed over 15 left-grasper peg-transfer trials.}
\label{tab:gripper_analysis}
\centering
\setlength{\tabcolsep}{10pt}
\begin{tabular}{lccc}
\toprule
\textbf{Modalities} & \textbf{IoU ($\uparrow$)} & \textbf{Acc ($\uparrow$)} & \textbf{Prec ($\uparrow$)} \\
\midrule
PSM--MTM     & 0.90 & 0.97 & 0.97 \\
EmHT--MTM    & \textbf{0.54} & \textbf{0.81} & \textbf{0.67} \\
EmHT--PSM    & \textbf{0.53} & \textbf{0.80} & \textbf{0.68} \\
\bottomrule
\end{tabular}
\end{table*}

\begin{figure}[htbp]
    \centering

    \begin{minipage}[htbp]{0.6\columnwidth}
        \centering
        \includegraphics[width=\linewidth]{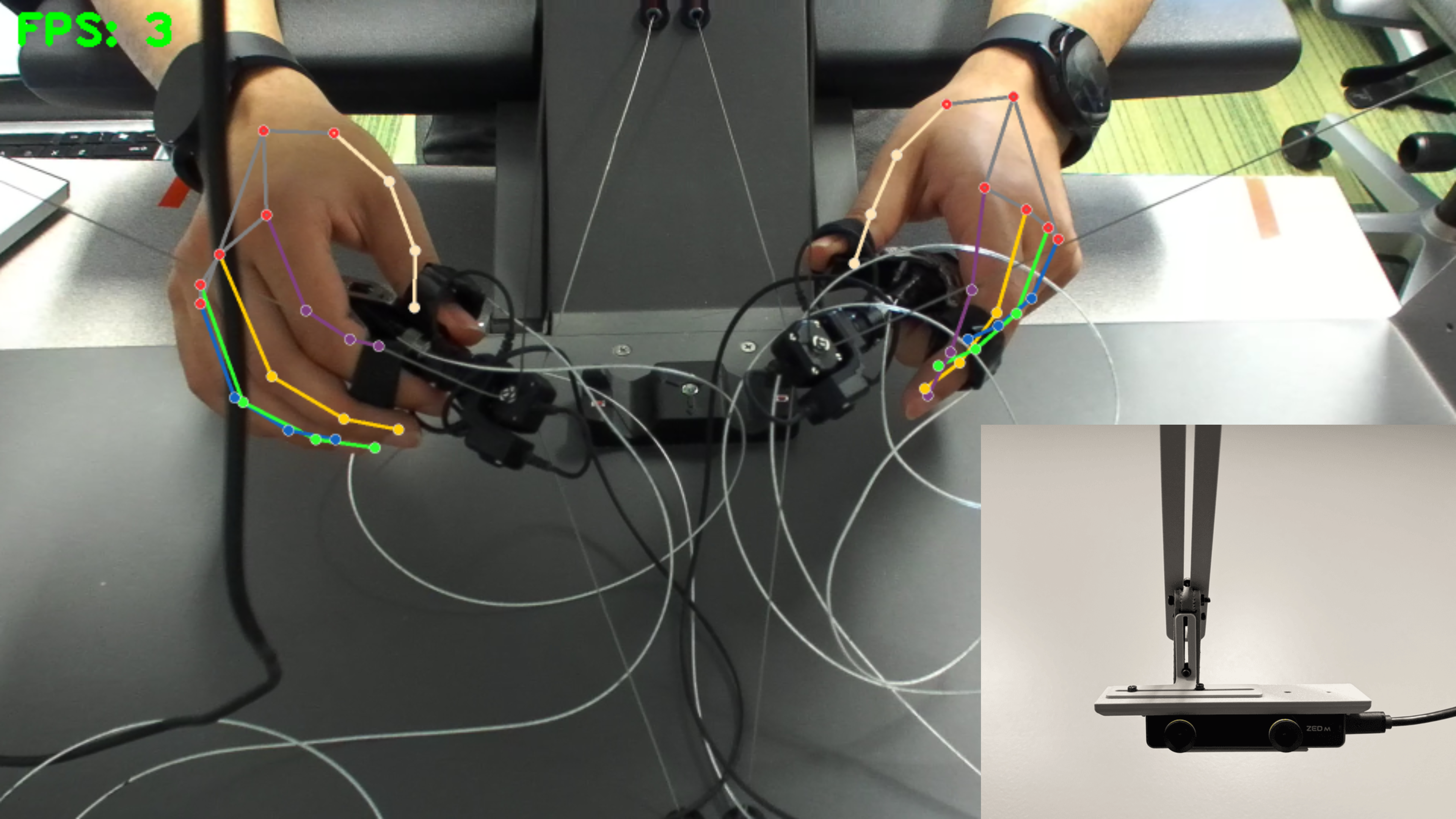}\\
        \textbf{(a)} The MTMs Workspace, Hand Keypoints and the Mounted RGB-D Camera.
    \end{minipage}

    \vspace{6pt}

    \begin{minipage}[htbp]{0.6\columnwidth}
        \centering
        \includegraphics[width=\linewidth]{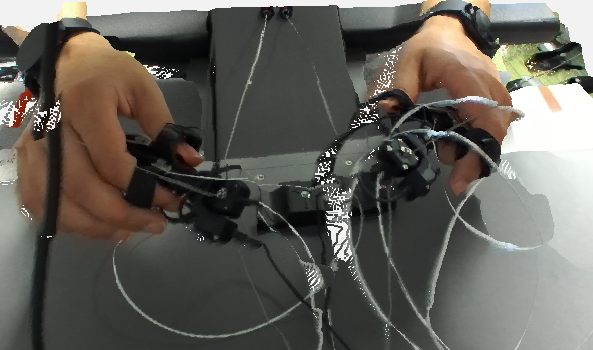}\\[-2pt]
        \textbf{(b)} The Point Cloud of Hands and MTMs.
    \end{minipage}

    \vspace{6pt}

    \begin{minipage}[htbp]{0.6\columnwidth}
        \centering
        \includegraphics[width=\linewidth]{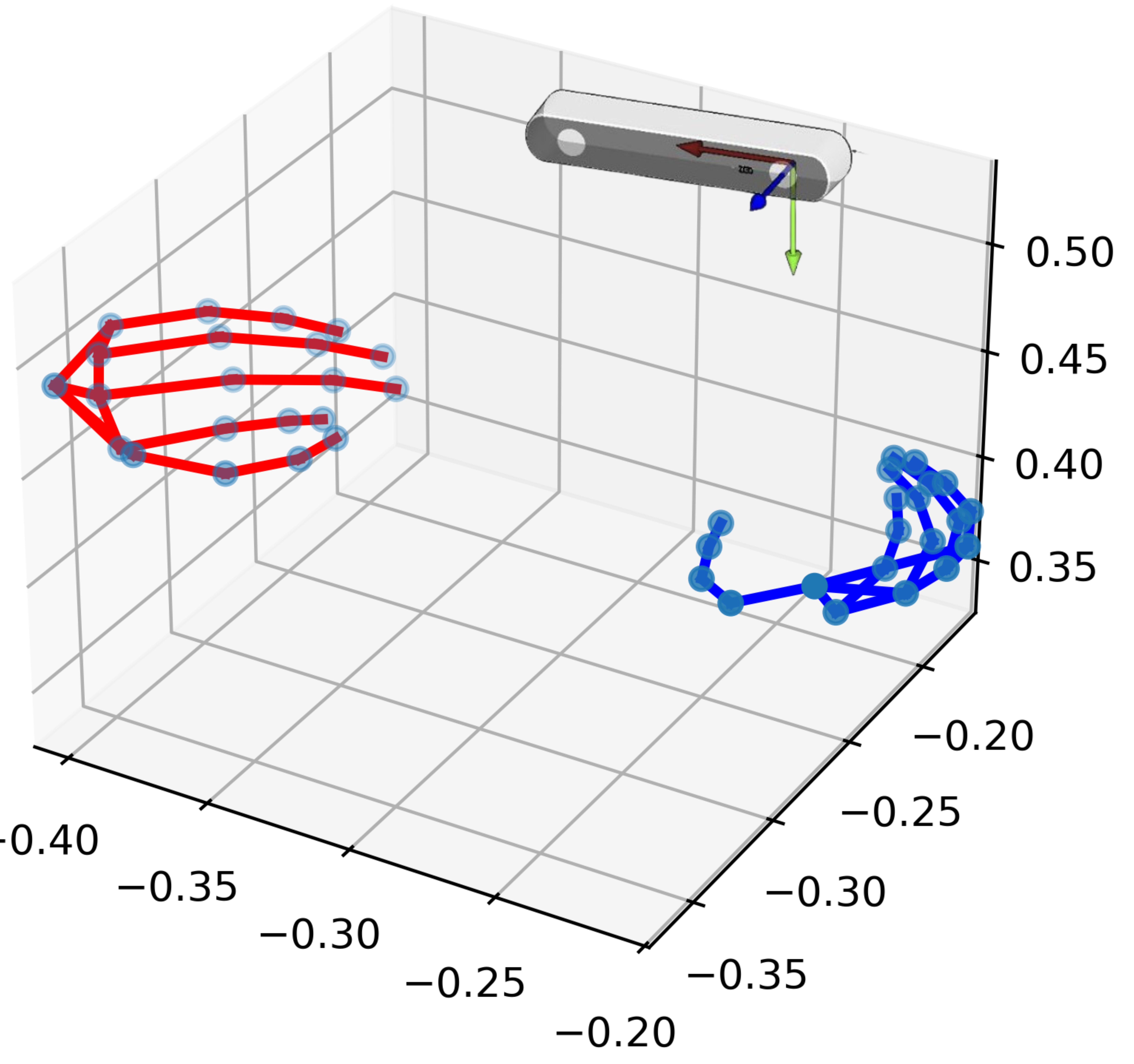}\\
        \textbf{(c)} The localized cartesian poses of hand keypoints in MTM workspace.
    \end{minipage}

    \caption{The surgeon hands and MTM workspace, captured by the RGB-D camera and processed extract (a) hand keypoints, (b) hands and MTMs point cloud and (c) localized hand keypoints in the workspace.}
    \label{fig:hand_rgbd}
\end{figure}

\subsection{Depth Camera-Based Hand Tracking}

\subsubsection{Keypoint Estimation and Transformation}
The hand and MTM pose estimation is performed using a two-step process described below:
\begin{enumerate}
    \item \textbf{Camera Calibration:} First, hand-eye calibration~\cite{hand_eye_calib} is performed between the MTMs (coordinate frame $M$) and the RGB-D camera (coordinate frame $C$) to obtain the rigid body transformation $\mathbf{T}^{M}_{C}$. This transformation, composed of a rotation matrix $\mathbf{R}^{M}_{C}$ and a translation vector $\mathbf{t}^{M}_{C}$, allows for representing all poses in the same MTM base frame. Calibration procedure is performed once per data collection session.
    \item \textbf{Keypoint Extraction and Transformation:} Hand keypoints are extracted from a camera view that is aligned with a calibrated depth map, using the Mediapipe's hand landmark detection library~\cite{mediapipe}. An example of such keypoints is shown in Figure \ref{fig:hand_rgbd} (a). Utilizing the aligned RGB and depth frames provided by the ZED software development kit, a point cloud is generated (see Figure \ref{fig:hand_rgbd} (b)), and each extracted hand keypoint $j$ is localized as a 3D position vector $\mathbf{p}_{C,j}$ in the camera's coordinate frame ($C$). Using the rigid body transformation from calibration, these keypoints are then transformed into the MTMs base frame ($M$) to obtain $\mathbf{p}_{M,j}$:
\begin{equation}
    \mathbf{p}_{M,j} = \mathbf{R}^{M}_{C} \, \mathbf{p}_{C,j} + \mathbf{t}^{M}_{C}
\end{equation}
The hand keypoints, represented in MTMs coordinate frame, are shown in Figure \ref{fig:hand_rgbd} (c). Finally, the estimated position of each MTM is taken to be the average positions of the thumb and index finger tips holding that MTM, and is referred to as HandKP.
\end{enumerate}

As an end-to-end alternative, we also experiment with a direct transformation from hand keypoints in camera frame to MTMs base coordinates without the calibration process using an MLP network (similar to the model used in Section \ref{sec:emht}) to assess the usability of the system for cameras with unknown calibration values. 
This is also evaluated using the same 15-fold LOTO-CV protocol described in Section \ref{sec:emht}.

\subsubsection{Keypoint Detection Accuracy}
We evaluated the \textit{HandKP} pipeline as a proxy for the MTMs trajectories over 15 Peg Transfer trials using the same 15-fold LOTO-CV protocol described in Section~\ref{sec:emht}. Two transformations were compared: the calibrated rigid camera-to-MTM transformation and the direct MLP mapping from camera-frame hand keypoints to MTM coordinates. Performance is reported using Mean Cosine Similarity (CoS), which measures trajectory-shape agreement, and Normalized Root Mean Square Error (NRMSE), which measures magnitude error relative to the dynamic range of the ground-truth signal. For an evaluated coordinate trajectory with ground truth $x_t$, prediction $\hat{x_t}$, and N evaluated frames, NRMSE is computed as:
\begin{equation}
\mathrm{NRMSE}(\%) =
100 \times
\frac{
\sqrt{\frac{1}{N}\sum_{t=1}^{N}\left(\hat{x_t}-{x_t}\right)^2}
}{
\max_{t}(x_t)-\min_{t}(x_t)
}
\end{equation}
Lower NRMSE indicates smaller normalized trajectory error.

RGB-D hand tracking is line-of-sight dependent, and missing HandKP samples occur when keypoints are not detected due to hand/MTM occlusions or when the surgeon's hands move outside the camera field of view. To preserve synchronized trajectories for downstream analysis, gaps of at most one second are filled using cubic spline interpolation, while longer gaps are forward-filled. Table~\ref{tab:hand_rgbd_include_missing} reports the full-trajectory evaluation after this gap handling, therefore reflecting both transformation accuracy and the effect of intermittent keypoint-detection failures.

\begin{table}[t]
\caption{Mean CoS and NRMSE (\%) RGB-D estimation of left MTM trajectory ($X$, $Y$, $Z$) using cross-validation over 15 Peg Transfer trials.}
\label{tab:hand_rgbd_include_missing}
\centering
\begingroup
\setlength{\tabcolsep}{2.5pt}
\renewcommand{\arraystretch}{0.95}
\resizebox{0.6\columnwidth}{!}{%
\begin{tabular}{@{}lcccccc@{}}
\toprule
\multirow{2}{*}{\textbf{Model}} &
\multicolumn{2}{c}{\textbf{X}} & \multicolumn{2}{c}{\textbf{Y}} & \multicolumn{2}{c}{\textbf{Z}} \\
\cmidrule(l){2-7}
  & CoS$\uparrow$ & NRMSE$\downarrow$ & CoS$\uparrow$ & NRMSE$\downarrow$ & CoS$\uparrow$ & NRMSE$\downarrow$ \\
\midrule
Rigid Transformation & 0.45 & 26.39 & 0.63 & 32.14 & 0.79 & 33.06 \\
MLP & 0.72 & 26.13 & 0.71 & 30.15 & 0.89 & 19.88 \\
\bottomrule
\end{tabular}%
}
\endgroup
\end{table}

With full trajectories, the MLP improves CoS on the X, Y and Z axes from 0.45, 0.63 and 0.79 to 0.72, 0.71 and 0.89 respectively, and substantially reduces Z-axis NRMSE from 33.06\% to 19.88\%. The MLP provides more consistent overall performance, particularly in reducing magnitude error.

To isolate transformation quality from missing-keypoint effects, we additionally evaluated only frames with successful hand detection and keypoint extraction; these results are reported in Table~\ref{tab:hand_rgbd_ignore_missing}. In this detected-frame setting, the MLP outperforms the rigid transformation across all axes, achieving CoS values of 0.91, 0.72, and 0.97 for X, Y, and Z, respectively, with NRMSE values between 10.96\% and 14.93\%.

\begin{table}[t]
\caption{Mean CoS and NRMSE (\%) RGB-D estimation of left MTM trajectory ($X$, $Y$, $Z$) using cross-validation over 15 Peg Transfer trials (Ignoring Missing Data).}
\label{tab:hand_rgbd_ignore_missing}
\centering
\begingroup
\setlength{\tabcolsep}{2.5pt}
\renewcommand{\arraystretch}{0.95}
\resizebox{0.6\columnwidth}{!}{%
\begin{tabular}{@{}lcccccc@{}}
\toprule
\multirow{2}{*}{\textbf{Model}} &
\multicolumn{2}{c}{\textbf{X}} & \multicolumn{2}{c}{\textbf{Y}} & \multicolumn{2}{c}{\textbf{Z}} \\
\cmidrule(l){2-7}
  & CoS$\uparrow$ & NRMSE$\downarrow$ & CoS$\uparrow$ & NRMSE$\downarrow$ & CoS$\uparrow$ & NRMSE$\downarrow$ \\ 
\midrule
Rigid Transformation & 0.81 & 16.48 & 0.67 & 14.75 & 0.93 & 16.00 \\
MLP & \textbf{0.91} & \textbf{10.96} & \textbf{0.72} & \textbf{14.19} & \textbf{0.97} & \textbf{14.93} \\
\bottomrule
\end{tabular}%
}
\endgroup
\end{table}

The improvement from Table~\ref{tab:hand_rgbd_include_missing} to Table~\ref{tab:hand_rgbd_ignore_missing} indicates that the primary limitation of RGB-D HandKP is intermittent keypoint availability caused by FoV and occlusion constraints, rather than the coordinate transformation itself. These results support depth-based hand tracking as a useful MTM-position proxy when detections are available, while also motivating wider-FoV or multi-camera setups for more robust continuous tracking. Based on its stronger detected-frame performance and lower full-trajectory errors, the MLP-transformed HandKP trajectory is used in the final accuracy evaluations and gesture-recognition experiments.


\subsection{Foot Pedal Sensing}

\begin{figure}
    \centering
    \includegraphics[width=0.8\linewidth]{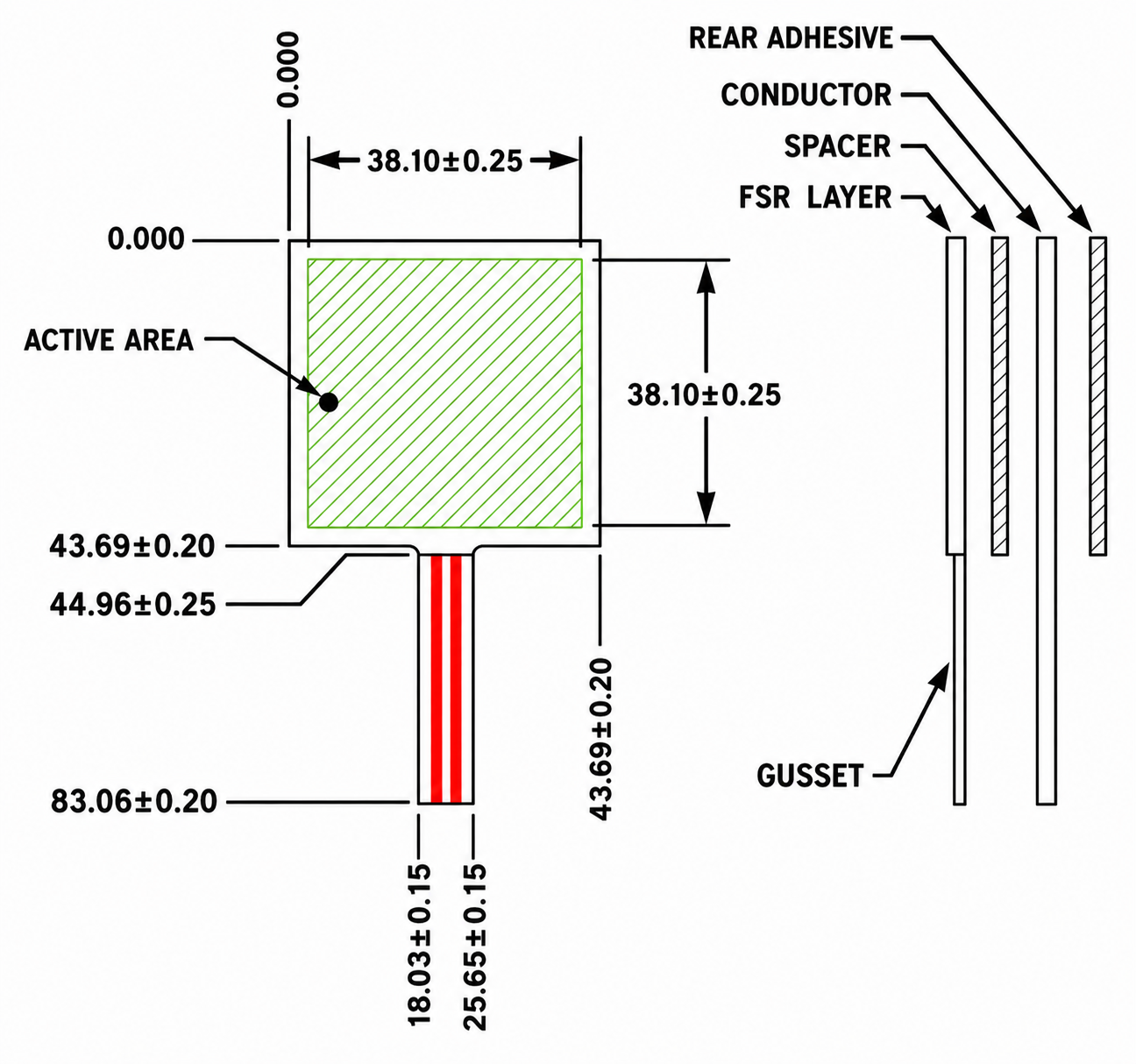}
    \caption{Dimensions of the FSR utilized for the PSS in mm~\cite{FSR-Data-Sheet}.}
    \label{fig:fsr-diag}
\end{figure}

Surgeon foot pedal interactions convey critical behavioral and operational cues during robot-assisted surgery, including instrument activation patterns, clutching behavior (hand motions decoupled from end-effector movement), and potential indicators of expertise. To capture these signals non-invasively, we developed a \textit{Foot Pedal Sensing (FPS)} module, a compact, low-cost, open source system built around an Arduino-class microcontroller and thin-film force-sensitive resistors (FSRs) (see Figure \ref{fig:fsr-diag}).

\begin{figure}
    \centering
    \includegraphics[width=0.8\linewidth]{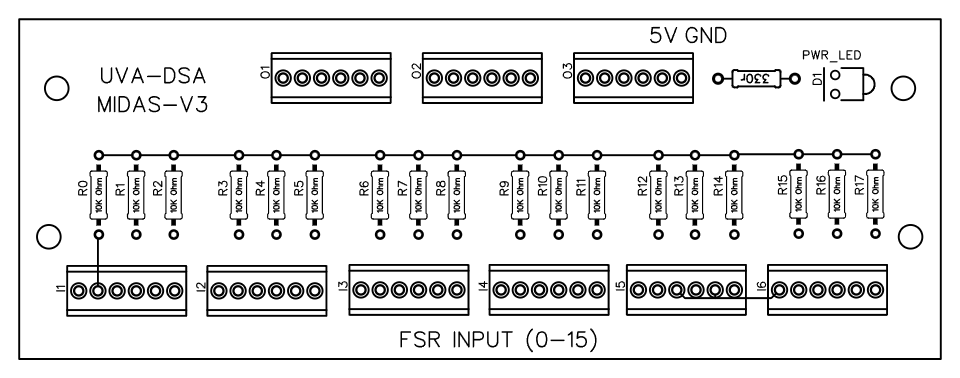}
    \caption{Custom designed PCB schematic for interfacing between microcontroller and FSR sensors.}
    \label{fig:pcb-schem}
\end{figure}

Each FSR is mounted in a voltage-divider with a known series resistor \(R\) using a custom designed PCB (see Figure \ref{fig:pcb-schem}). Under applied pressure, the FSR resistance \(R_{\mathrm{FSR}}\) decreases, increasing the divider output \(V_{\mathrm{out}}\) in Eq. (\ref{eq:pss_vout}), which is sampled by the microcontroller’s analog-to-digital converter.

\begin{equation}
  V_{\mathrm{out}} = \frac{R}{R + R_{\mathrm{FSR}}}\,V_{CC}.
  \label{eq:pss_vout}
\end{equation}

A brief per-pedal calibration establishes voltage thresholds for robust binary actuation detection at the start of the sensing.
Following the initial calibration, we further refine this value through data-driven optimization. Using a subset of labeled ground-truth pedal data, we perform a hyperparameter search over possible voltage thresholds to identify the value that maximizes detection performance. The resulting optimal threshold is then applied to the dataset to evaluate overall PSS accuracy.

Sensors are affixed directly to the surgeon's console pedal surfaces, enabling continuous monitoring without modifying console hardware or perceptibly altering tactile feedback. 
PSS supports reading up to 9 pedals and records each pedal state along with time-stamped \(V_{\mathrm{out}}\) values for detailed analysis.

Signals are time-stamped and logged by the MiDAS acquisition system at a specified sampling rate, supporting seamless integration into existing workflows while preserving natural surgeon interaction and providing a high-value stream for behavioral and skill analysis.



\section{Downstream Task Evaluation: Gesture Recognition}\label{benchmark}

We evaluate the effectiveness of MiDAS for downstream gesture recognition using both the collected MiDAS datasets and additional established benchmarks. Experiments on MiDAS datasets (Raven-II Peg Transfer and da Vinci Xi Suturing) assess the performance of non-invasive sensing modalities in realistic settings. We employ state-of-the-art (SOTA) models for our benchmarks, consisting of a transformer-based model, MTRSAP~\cite{weerasinghe2024multimodal}, and a convolutional neural network (CNN)-based model, MSTCN++~\cite{farha2019ms}. In addition, to contextualize these results and examine the impact of dataset scale, we include evaluations on larger external datasets (DESK~\cite{madapana2019desk}, SAR-RARP50~\cite{psychogyios2023sar}, and SAIS~\cite{Kiyasseh2023SAIS}) commonly used in prior work. These external datasets are used solely for comparative benchmarking and are not part of the MiDAS data collection.

\subsection{Raven-II Peg Transfer Ablation Study}

Table~\ref{tab:abl_modality_appendix} summarizes the Peg Transfer ablation study on key kinematic features and robot states, including position, orientation, velocity, grasper state, and clutch state. Due to the MiDAS Raven-II dataset being moderate in scale, we also include DESK Peg Transfer results as larger-dataset comparison baselines. These rows are intended to contextualize the effect of dataset size and should not be interpreted as part of the MiDAS Raven-II feature ablation.

\begin{table*}[htbp]
\centering
\resizebox{0.8\textwidth}{!}{%
\begin{tabular}{@{}llllcccc@{}}
\toprule
\textbf{Dataset} & \textbf{Model} & \textbf{Modality} & \textbf{Features} &
\textbf{Acc} & \textbf{F1} & \textbf{Prec.} & \textbf{Rec.} \\
\midrule

\multirow{14}{*}{\centering\arraybackslash\shortstack[l]{\textbf{Raven-II}\\[1pt]Peg Transfer\\[0pt](15 trials)}}
 & \multirow{8}{*}{\textbf{MTRSAP}}
   & PSM        & pos.,  ori., grasp. 
                & 0.72 & 0.75 & 0.72 & 0.71 \\
 & & PSM-NG     & pos.,  ori 
                & 0.61 & 0.58 & 0.61 & 0.60 \\
& & \textbf{PSM}        & \textbf{pos.,  ori., vel., grasp.} 
                &\textbf{0.88}  & \textbf{0.87} & \textbf{0.88}  & \textbf{0.87} \\
\cmidrule(l){3-8}

 & & MTM        & pos., ori., grasp. 
                & 0.77 & 0.78 & 0.77 & 0.76 \\
 & & MTM-NG     & pos., ori. 
                & 0.48 & 0.47 & 0.49 & 0.46 \\
& & MTM        & pos., ori., vel., grasp. 
                &  0.87 & 0.87  & 0.90  & 0.86  \\
\cmidrule(l){3-8}
 & & EmHT & pos. 
                & 0.68  & 0.67  & 0.68  & 0.67  \\
 & & EmHT & pos., ori. 
                & 0.73 & 0.75 & 0.73 & 0.74 \\


 & & EmHT & pos., vel. 
                & 0.84 & 0.84 & 0.85 & 0.84 \\

 & & \textbf{EmHT} & \textbf{pos., ori., vel.} 
                & \textbf{0.87} & \textbf{0.86} & \textbf{0.87} & \textbf{0.86} \\
 & & EmHT-NC & pos., ori., vel., no-clutch 
                & 0.78 & 0.77 & 0.79 & 0.77 \\
\cmidrule(l){3-8}

 & & Image$\dagger$ & RGB (R) 
                & 0.39 & 0.42 & 0.39 & 0.39 \\
 & & Image$\dagger$ & RGB (D) 
                & 0.48 & 0.48 & 0.48 & 0.48 \\
 \cmidrule(l){3-8}


 & & HandKP    & pos., vel. 
                & 0.40 & 0.38 & 0.41 & 0.39 \\
                
  & & HandKP, Image    & pos., vel., RGB (R)  
                & 0.25  & 0.16  &  0.19 &  0.24 \\    
  & & HandKP, Image    & pos., vel., RGB (D)  
                &  0.38 & 0.33  & 0.36  & 0.37  \\

\cmidrule(l){2-8}

 & \multirow{6}{*}{\textbf{MS-TCN++}}
   & PSM        & pos., ori., grasp. 
                & 0.74 & 0.76 & 0.75 & 0.74 \\
 &   & PSM        & pos., ori., vel., grasp. 
                & 0.85 & 0.84 & 0.86 & 0.86 \\
\cmidrule(l){3-8}

 & & MTM & pos., ori., grasp. 
                & 0.76 & 0.78 & 0.76 & 0.74 \\
 & & \textbf{MTM} & \textbf{pos., ori., vel., grasp.} 
                & \textbf{0.86} & \textbf{0.86} & \textbf{0.86} & \textbf{0.87} \\

\cmidrule(l){3-8}

 & & EmHT & pos. 
                & 0.55 & 0.53 & 0.55 & 0.56 \\
 & & EmHT & pos., ori 
                & 0.65 & 0.63 & 0.68 & 0.68 \\

 & & EmHT & pos., vel. 
                & 0.74 & 0.74 & 0.77 & 0.76 \\

 & & \textbf{EmHT} & \textbf{pos., ori., vel.} 
                & \textbf{0.79} & \textbf{0.78} & \textbf{0.80} & \textbf{0.81} \\

\cmidrule(l){3-8}
 
 & & Image$\dagger$  & RGB (R) 
                & 0.44 & 0.50 & 0.45 & 0.43 \\
 & & Image$\dagger$ & RGB (D) 
                & 0.40 & 0.47 & 0.41 & 0.44 \\

\cmidrule(l){3-8}
                
 & & HandKP    & pos., vel.
                & 0.33 & 0.29 & 0.27 & 0.37 \\
                
   & & HandKP, Image  & pos., vel., RGB (R)
                         & 0.33  & 0.32  & 0.32  &  0.36  \\                   
  & & HandKP, Image  & pos., vel., RGB (D)
                         & 0.49  & 0.51  & 0.49  &  0.46 \\           

\midrule

\multirow{4}{*}[-0.5ex]{\centering\arraybackslash\shortstack[l]{\textbf{DESK}\\[1pt]Peg Transfer\\[0pt](48 trials)}}
 & \multirow{2}{*}{\textbf{*MTRSAP}}
   & PSM        & pos., vel., ori., grasp. 
                & 0.91 & 0.91 & 0.92 & 0.92 \\
 & & Image  & RGB (R) 
                & 0.76 & 0.77 & 0.77 & 0.76 \\
\cmidrule(l){2-8}
 & \multirow{2}{*}{\textbf{*MS-TCN++}}
   & PSM        & pos., vel., ori., grasp. 
                & 0.92 & 0.93 & 0.92 & 0.92 \\
 & & Image  & RGB (R) 
                & 0.73 & 0.77 & 0.75 & 0.75 \\
\bottomrule
\end{tabular}%
}
\caption{Peg Transfer gesture recognition ablation using different modalities. 
\textbf{Bold} values indicate the best-performing feature configuration for selected modalities within each model.
\(\dagger\) Image-only modalities underperform due to limited data. * Same baselines evaluated on a larger dataset for comparison. \textit{Abbreviations: pos.=position; vel.=velocity; ori.=orientation; grasp.=grasper angle; NG=No Grasper State Information; NC=No Clutch State Information; (R)=ResNet-50; (D)=DINOv2.}}
\label{tab:abl_modality_appendix}
\end{table*}

\subsubsection{Kinematic Feature and Robot-State Ablations}

For a comprehensive evaluation of the impact of different features on gesture recognition performance, we performed an ablation study on key kinematic features and robot states including position, orientation, velocity, grasper state, and clutch state. Across the Raven-II rows in Table~\ref{tab:abl_modality_appendix}, non-invasive EmHT with position, orientation, and velocity achieves performance close to internal robot kinematics, indicating that external hand-motion sensing can provide a strong proxy for downstream gesture recognition.

We first incorporate clutch state information, which marks periods when surgeon hand motion is decoupled from PSM movement. Excluding these clutching windows during training and inference removes kinematically irrelevant segments and leads to notable performance gains for EmHT; F1 improves from 0.77 to 0.86, demonstrating that clutch state is a valuable auxiliary signal. Given these benefits, clutch state is included for all remaining modalities.

We then evaluated the significance of grasper (jaw) state within the kinematic stream in both PSM and MTM. Removing this signal, an indicator of the surgeon's pinching action, caused \textbf{macro-F1 to drop significantly from 0.78 to 0.47 in MTM performance and 0.75 to 0.58 in PSM performance}, highlighting its discriminative value. Notably, EmHT (two fingers per manipulator) indirectly captures the pinching motion, providing a proxy for grasper state and partially mitigating the loss when explicit grasper signals are unavailable, as discussed previously in the EmHT correlation analysis.

We further evaluate feature ablations on position, orientation, and velocity, and find that incorporating velocity consistently yields substantial gains across kinematic modalities, highlighting its importance for gesture recognition.

\subsubsection{Visual Feature Adaptation and Pretraining Strategies}

In Raven-II Peg Transfer experiments, adding vision features (ResNet-50~\cite{wang2018deep}, DINOv2~\cite{oquab2023dinov2}) initially failed to surpass kinematic baselines. We attribute this to the low-data regime: high-dimensional encoders trained on few clips overfit appearance rather than gesture-discriminative motion; the supervision-to-dimension mismatch (clip-level labels vs.\ millions of visual DoF) worsens this, and ImageNet-to-surgery domain shift~\cite{deng2009imagenet} (illumination, background, and surgical instruments) further limits transfer.

To mitigate the low-data regime, we first pretrained ResNet-50 on the in-domain DESK dataset~\cite{madapana2019desk}, which yielded a 15-30\% relative improvement, demonstrating the clear benefit of domain-adaptive pretraining. Replacing ResNet-50 with DINOv2 features provided further gains, consistent with the strength of large-scale self-supervised vision transformers that capture more semantically expressive and globally contextualized representations. Overall, our results highlight that (i) domain-adaptive pretraining is particularly important for Convolutional Neural Network (CNN) backbones, and (ii) foundation model features such as DINOv2 offer an even stronger initialization under limited supervision, especially when combined with motion cues.

Consistent with the RGB-D hand-tracking analysis, HandKP-based recognition remains limited by intermittent keypoint availability caused by field-of-view and occlusion constraints. Therefore, although HandKP provides a useful non-invasive motion cue, EmHT provides the most reliable non-invasive motion stream for Peg Transfer gesture recognition in this setting.

\subsection{da Vinci Xi Suturing Ablation Study}

Table~\ref{tab:abl_modality_suturing_louo_appendix} reports suturing gesture recognition on the MiDAS da Vinci Xi dataset using different modalities. We also include SAR-RARP50~\cite{psychogyios2023sar} and SAIS~\cite{Kiyasseh2023SAIS} results as larger-dataset external comparison baselines to contextualize performance under different dataset scales; these rows are not part of the MiDAS da Vinci Xi modality ablation.

\begin{table*}[htbp]
\centering
\resizebox{1\textwidth}{!}{%
\begin{tabular}{@{}llllcccc@{}}
\toprule
\textbf{Dataset} & \textbf{Model} & \textbf{Modality} & \textbf{Features} &
\textbf{Acc} & \textbf{F1} & \textbf{Prec.} & \textbf{Rec.} \\
\midrule

\multirow{11}{*}{\centering\arraybackslash\shortstack[l]{\textbf{da Vinci Xi}\\[1pt]Suturing\\[0pt](17 trials)}}
 & \multirow{6}{*}{\textbf{MTRSAP}}
   


 & Image           & RGB (R) 
                         & 0.55 & 0.57 & 0.57 & 0.56 \\
 & & Image           & RGB (D) 
                         & 0.69 & 0.67 & 0.65 & 0.65 \\
\cmidrule(l){3-8}

 & & EmHT                & pos., ori., vel.
                         & 0.64 & 0.60 & 0.63 & 0.61 \\
  & & EmHT, Image
                         & pos., ori., vel., RGB (R) 
                         & 0.60  &  0.57 & 0.63  & 0.57  \\

  & & \textbf{EmHT, Image} 
                         & \textbf{pos., ori., vel., RGB (D)} 
                         & \textbf{0.71} & \textbf{0.70} & \textbf{0.70} & \textbf{0.69} \\
\cmidrule(l){3-8}


  & & HandKP    & pos., vel.
                & 0.40 & 0.35 & 0.37 & 0.36 \\
                
  & & HandKP, Image & pos., vel., RGB (R) 
                         & 0.50  &  0.46 & 0.49  & 0.47  \\
                         
 & & HandKP, Image & pos., vel., RGB (D) 
                         & 0.58 & 0.58 & 0.58 & 0.57 \\

\cmidrule(l){2-8}
 & \multirow{5}{*}{\textbf{MS-TCN++}}

  & Image          & RGB (R)
                         & 0.53 & 0.44 & 0.44 & 0.43 \\
 & & Image          & RGB (D)
                         & 0.57 & 0.57 & 0.53 & 0.51 \\
\cmidrule(l){3-8}

  &  & EmHT                & pos., ori., vel. 
                         & 0.57 & 0.50 & 0.50 & 0.53 \\
 & & EmHT, Image 
                         & pos., ori., vel., RGB (R)
                         & 0.58  & 0.51 & 0.51  & 0.52 \\

 & & \textbf{EmHT, Image} 
                         & \textbf{pos., ori., vel., RGB (D)}
                         & \textbf{0.63} & \textbf{0.56} & \textbf{0.64} & \textbf{0.63} \\
\cmidrule(l){3-8}
                         
   & & HandKP    & pos., vel.
                &  0.43 & 0.36  & 0.39  & 0.37  \\

& & HandKP, Image  & pos., vel., RGB (R)
                         & 0.48 & 0.42  & 0.46   & 0.45   \\
  & & HandKP, Image  & pos., vel., RGB (D)
                         & 0.58 & 0.54  &  0.58  &  0.54  \\

\midrule

\multirow{2}{*}[0.3ex]{\centering\arraybackslash\shortstack[l]{\textbf{SAR-RARP50}\\[1pt]Suturing (50 trials)}} 
 & \textbf{ASFormer~\cite{yi2021asformer}} & Image 
   & RGB 
   & $0.82^{\dagger}$ & -- & -- & -- \\
 & \textbf{MViT~\cite{fan2021multiscale}} & Image 
   & RGB 
   & $0.79^{\dagger}$ & -- & -- & -- \\
\midrule

\multirow{2}{*}[0.2ex]{\centering\arraybackslash\shortstack[l]{\textbf{SAIS}\\[1pt]Suturing (78 trials)}} 
 & \multirow{2}{*}{\shortstack[l]{\textbf{Vision}\\\textbf{Transformer}~\cite{Kiyasseh2023SAIS}}}
   & Image & RGB 
   & -- & $0.50^{\dagger}$ & -- & -- \\
 & & \multicolumn{6}{l}{} \\
\bottomrule
\end{tabular}%
}
\caption{
Suturing gesture recognition on the MiDAS da Vinci Xi dataset using different modalities, with SAR-RARP50 and SAIS included as larger baseline datasets for comparison. Best results per model on the MiDAS da Vinci Xi dataset are shown in \textbf{bold}. External dataset rows are included only for contextual comparison and are not part of the MiDAS da Vinci Xi modality ablation.
\textit{Abbreviations: pos.=position; vel.=velocity; ori.=orientation; (R)=ResNet-50; (D)=DINOv2; $\dagger$=As reported or estimated from published results.}}
\label{tab:abl_modality_suturing_louo_appendix}
\end{table*}

\subsubsection{Aggregate Modality Trends}

Table~\ref{tab:abl_modality_suturing_louo_appendix} highlights several key modality trends on the MiDAS da Vinci Xi Suturing dataset. First, \textbf{image-only} models show moderate performance, with DINOv2 features outperforming ResNet-50 (Acc: 0.69 vs.\ 0.55 for MTRSAP), reflecting the benefit of foundation-model representations under low-data conditions. Second, \textbf{EmHT alone} performs competitively (F1: 0.60 with MTRSAP), indicating that external hand-motion trajectories capture meaningful cues for suturing despite the complexity of real surgical tissue interactions. Third, \textbf{fusing EmHT with visual features} yields the strongest results (Acc: 0.71, F1: 0.70 for MTRSAP), demonstrating that motion dynamics and visual context are complementary for recognizing fine-grained suturing gestures.

In contrast, \textbf{HandKP} features derived from RGB-D hand keypoint tracking perform notably worse in isolation (F1: 0.35), but offer moderate gains when fused with image features. This is expected, as keypoints are occasionally missed when the hands leave the camera's field of view, leading to fragmentary motion information.

\subsubsection{Per-Class and Error-Pattern Analysis}

\begin{figure}[htbp]
\centering{\includegraphics[width=0.6\columnwidth]{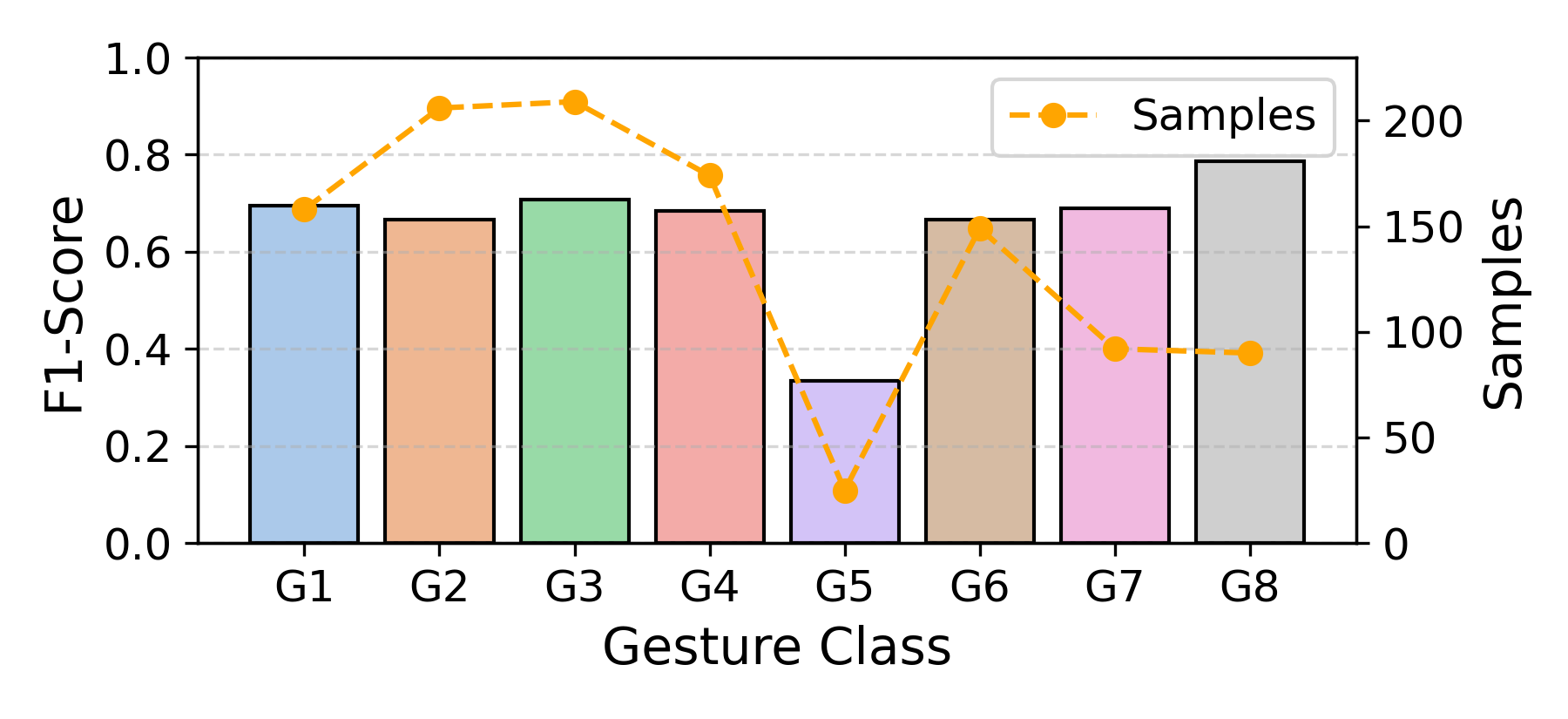}}
\caption{Per-class F1-score and class distribution for the MiDAS da Vinci Xi Suturing dataset using the best-performing MTRSAP configuration with fused EmHT and DINOv2 image features.}
\label{fig:bootcamp_suturing_perclass}
\end{figure}

\begin{figure}[!h]
\centering{\includegraphics[width=0.6\columnwidth]{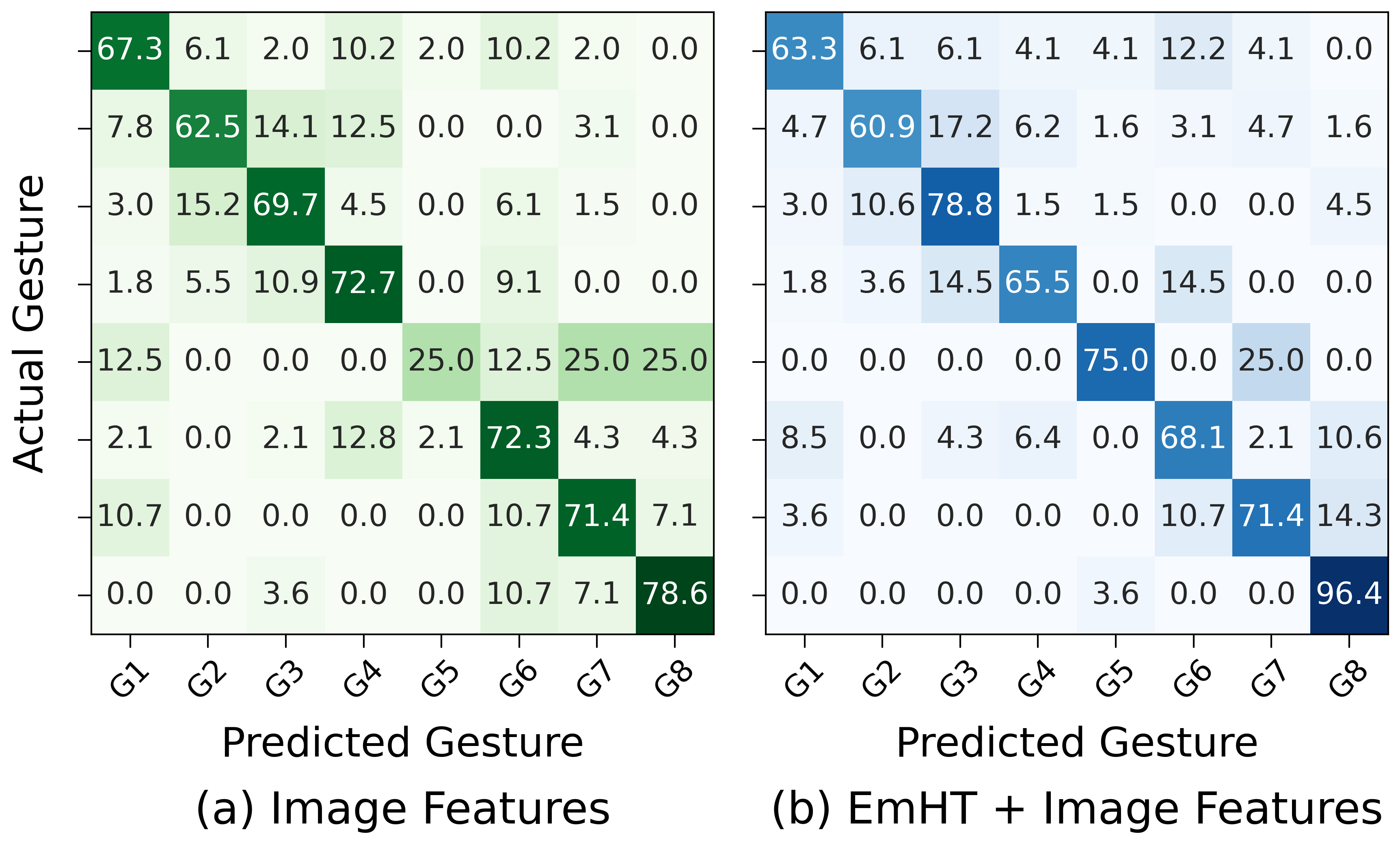}}
\caption{Confusion matrices for da Vinci Xi suturing gesture recognition using MTRSAP: (a) performance using image-only features and (b) performance using fused image and EmHT features.}
\label{fig:confusion_matrix}
\end{figure}

For the Suturing task, we report per-class gesture recognition performance alongside class distribution statistics to better characterize model performance and generalization across gestures (Figure~\ref{fig:bootcamp_suturing_perclass}). The best-performing model, MTRSAP~\cite{weerasinghe2024multimodal}, generalizes well over the gesture classes except for G5-Reach Suture, which has a significantly low number of samples for training. Moreover, Figure~\ref{fig:confusion_matrix} presents confusion matrices for visualization of how the best-performing models recognize gestures and which gestures are harder to distinguish. For example, on the da Vinci dataset, fusing EmHT with image features produces clear gains for gestures emphasizing fine needle control (G8-Square Knot and Cinch, G3-Push Needle Through Tissue) and imbalanced classes with fewer samples (G5), but lowers accuracy for visually similar pulling motions (G4-Pull Needle out of Tissue, G6-Pull Suture) as the hand movements are closely aligned kinematically.

Overall, the Suturing ablations show that (i) \textbf{DINOv2 image features} provide the strongest visual baseline, (ii) \textbf{EmHT} is an effective non-invasive proxy for robot-side motion, and (iii) \textbf{multimodal fusion} yields the most robust performance across gesture types, especially under class imbalance and variable tissue interactions.

\bibliographystyle{unsrtnat}
\bibliography{references}  

\end{document}